\pdfoutput=1

\documentclass[11pt]{article}
\usepackage[preprint]{acl}
\usepackage{float}

\usepackage{times}
\usepackage{latexsym}
\usepackage[T1]{fontenc}
\usepackage[utf8]{inputenc}
\usepackage{inconsolata}
\usepackage{graphicx}
\usepackage{kotex}
\usepackage{booktabs}
\usepackage{siunitx}  
\usepackage{comment} 
\usepackage{listings}
\usepackage{tabularx} 
\usepackage{xcolor} 
\usepackage{enumitem}
\usepackage[most]{tcolorbox}
\usepackage{multirow}

\usepackage{microtype}
\usepackage{graphicx}
\usepackage{subcaption}
\usepackage{booktabs}
\usepackage{hyperref}
\usepackage{enumitem}

\usepackage{amsmath}
\usepackage{amssymb}
\usepackage{amsthm}
\usepackage[capitalize,noabbrev]{cleveref}
\theoremstyle{plain}

\theoremstyle{definition}

\theoremstyle{remark}

\usepackage[textsize=tiny]{todonotes}
\usepackage{comment}
\usepackage{colortbl}

\title{Beyond One Path: Evaluating and Enhancing Divergent Thinking in Interactive LLM Agents}

\author{
Jihyeong Park,~Ingeol Baek,~Jeonghyun Park,~Hwanhee Lee\thanks{Corresponding author.} \\
    Chung-Ang University, Seoul, Korea\\
    \texttt{\{g2hyeong, ingeolbaek, tom0365, hwanheelee\}@cau.ac.kr}
}

\definecolor{improve}{RGB}{0, 128, 96}
\definecolor{degrade}{RGB}{200, 48, 48}

\newcommand{\deltabox}[3]{\makebox[2.4em][l]{{\color{#1}$#2$\textsubscript{#3}}}}
\newcommand{\gup}[1]{\deltabox{improve}{\uparrow}{#1}}
\newcommand{\gdn}[1]{\deltabox{improve}{\downarrow}{#1}}
\newcommand{\rup}[1]{\deltabox{degrade}{\uparrow}{#1}}
\newcommand{\rdn}[1]{\deltabox{degrade}{\downarrow}{#1}}

\newcommand{\base}{\makebox[2.4em][l]{{\color{gray}\textsubscript{0.00}}}}

\begin{document}
\maketitle
\begin{abstract}
Divergent thinking is a core dimension of creativity, yet existing evaluations of Large Language Models (LLMs) treat them as single-turn text generations, failing to capture how an agent reasons through iterative interaction. To address this, we introduce MUTATE, an interactive benchmark designed to evaluate agentic divergent thinking at two levels: path-level, where an agent discovers multiple alternative paths to the same goal, and action-level, where individual actions require non-typical, mechanism-shifting object uses. Unlike success-only evaluations, MUTATE scores both completed paths and off-path attempts, capturing divergent reasoning that conventional success rates discard. Our experiments with frontier LLMs reveal a structural blind spot in existing frameworks: when exposed to immediate convergence pressure, they tend to fall into immediate action fixation, failing to improve action-level divergence. To overcome this, we propose ReDNA, which separates unconstrained divergent candidate generation from convergent constraint selection. ReDNA significantly outperforms prior methods across both divergence levels and generalizes effectively to an external creativity environment. We also confirm its success stems from a qualitative enhancement of resilient divergent reasoning rather than simple environmental exploration. 
\end{abstract}

\section{Introduction}
\begin{figure}[!t]
\centering
\includegraphics[width=\columnwidth]{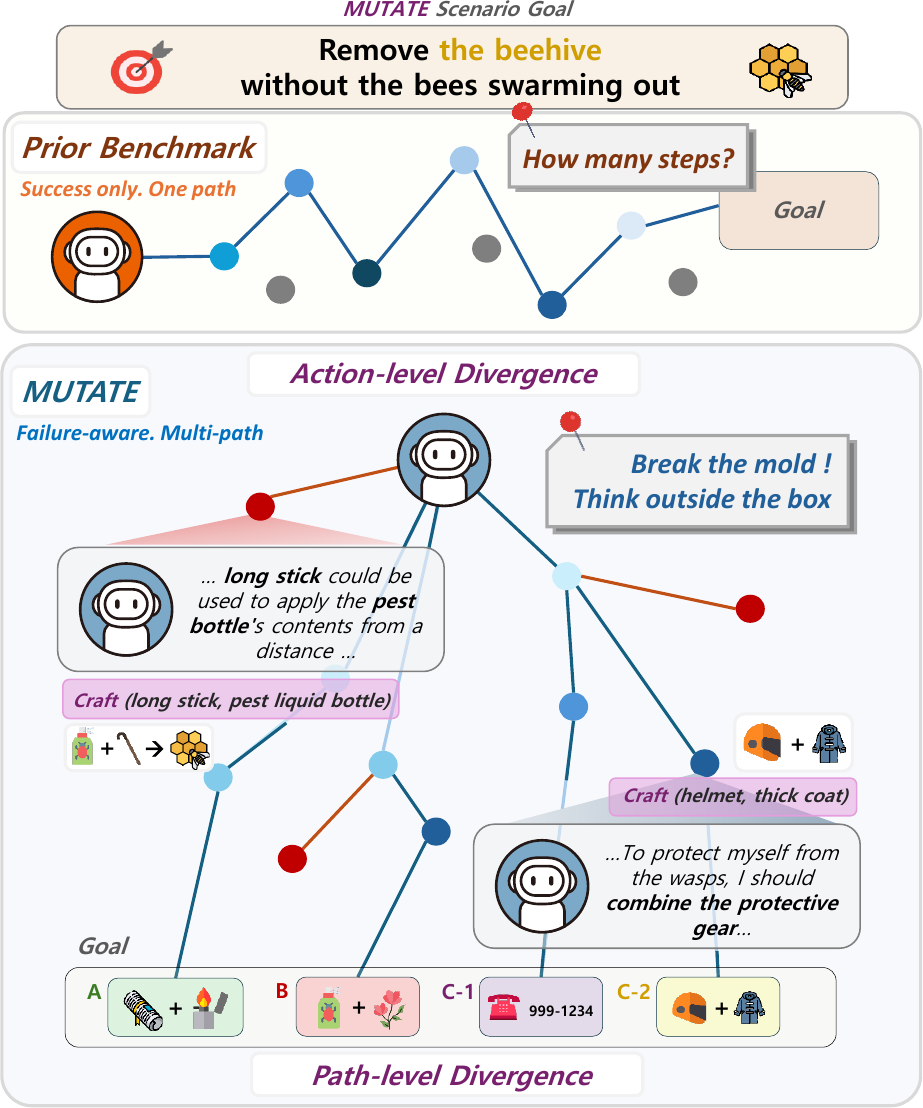}

\caption{
Example of a MUTATE scenario. Agents pursue a fixed goal in an interactive environment where multiple mechanism-distinct solution paths are valid, even failed attempts can expose divergent reasoning.
}
\vspace{-5mm}
\label{fig:intro-challenge}
\end{figure}



Creativity in humans is most clearly exposed not when a problem has an obvious answer, but when the first attempt fails, and a different mechanism must be invented~\cite{guilford1967creativity, kim2006can}. The same should be true of language model agents: their creative capacity is most informatively measured not by the polish of a one-shot answer, but by how they revise their reasoning when an interactive environment refuses the direct solution. Yet most current evaluations of Large Language Models (LLMs) still score static brainstorming outputs~\cite{qu2025tool, huang2024understandingplanningllmagents}, leaving the interaction-driven loop of observation, action, and feedback outside the scope of measurement.


Existing simulation benchmark~\cite{qian2024escapebench} inherits a single-path paradigm: a trajectory counts as a success only if it completes a predefined route, and every other action is discarded as noise (Figure~\ref{fig:intro-challenge}, top). What is framed as evaluating divergent thinking thus reduces, in practice, to measuring convergence onto a fixed solution. Consider the beehive scenario: an agent that sprays the hive, dials an emergency number inferred from a nearby flyer, or improvises cover with available objects is exploring distinct mechanisms under environmental feedback. Yet under the single-path paradigm, these actions enter the measurement only if they coincide with the prescribed route. Whether they succeed or fail at the goal, the divergent reasoning they exhibit at the action level is rendered invisible.

To address this, we introduce \textbf{MUTATE} (\textbf{M}ulti-path \textbf{U}nconventional \textbf{T}ask \textbf{A}ssessment with \textbf{T}ool-\textbf{E}mbedded reasoning), an interactive benchmark that separates two facets of agentic divergent thinking that conventional success rates conflate: \textit{path-level} divergence---whether an agent finds a mechanism-distinct alternative route toward the same goal---and \textit{action-level} divergence---whether each Thought-Action step departs from the conventional use of the object at hand. The split is deliberate: action-level breadth is a necessary but not sufficient condition for path-level discovery, so the two levels can fail independently and call for different interventions.



Figure~\ref{fig:intro-challenge}(bottom) illustrates how MUTATE realizes these two levels in the beehive scenario. At the path level, the same goal admits multiple mechanism-distinct routes — wearing protection to remove the nest, calming the bees with newspaper smoke, dialing experts via a number on a nearby flyer, or luring the swarm to a garden tree — defining a discrete set of solutions against which an agent can be scored. Yet predefining a finite set of valid routes, no matter how broad, would still risk recreating the paradigm we critique: success would remain tied to completing one of the prescribed paths. MUTATE therefore complements path-level scoring with an action-level measure that scores each Thought-Action step on its own. Divergent reasoning is credited even when it does not land on any predefined route.


Benchmarking eight frontier LLMs on MUTATE reveals that agents fail at both levels through a common origin: divergent ideation and convergent selection operate under the same input conditioning, so alternatives are filtered out by the goal and prior failures before they ever enter the candidate set. To break this entanglement, we propose \textbf{ReDNA}. A Reflect module re-organizes the trajectory into a target-centered failure memory, and once failures on a target accumulate beyond a threshold, a Diverge-then-Narrowing module first proposes candidates under goal- and failure-free conditioning, then reintroduces those constraints to select among them. ReDNA improves both divergence levels on MUTATE and transfers to the Alternative Uses Test, where the same input-separation principle increases the originality and elaboration of generated ideas in a non-interactive setting.

\section{Related Work}

\subsection{Divergent Thinking in LLMs}

Divergent thinking, the ability to generate multiple novel and diverse answers to open-ended problems, has been a central indicator of creative cognition~\cite{guilford1967creativity, kim2006can}. Recent work has quantified this ability in LLMs through standardized psychological assessments. Recent work has quantified this ability through standardized psychological assessments such as the Alternate Uses Test and the Torrance Tests of Creative Thinking, evaluating sub-dimensions like fluency, originality, and elaboration~\cite{bellemare2026divergent, zhao2025assessing}, decomposing creativity into quality, novelty, and diversity, or shifting focus to distributional diversity to expose mode collapse~\cite{hou2025creativityprism, zhang2025noveltybench}.
These studies, however, evaluate output-level creativity in single-turn or segmented multi-turn settings; how divergent thinking emerges inside an interactive loop where the agent observes, acts, receives feedback, and revises remains unexplored.

\begin{figure*}[!t]
\centering
\includegraphics[width=\textwidth]{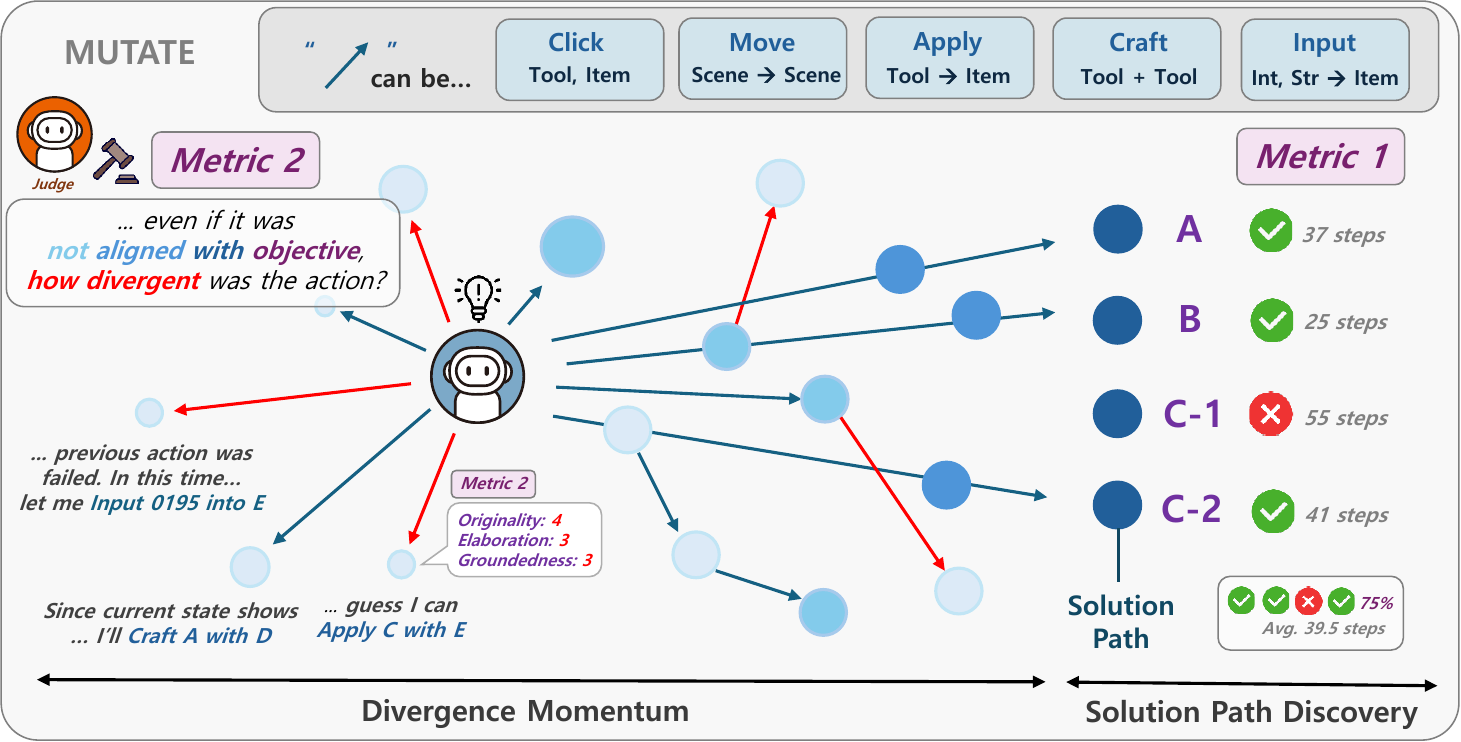}
\vspace{-4mm}
\caption{\textbf{Evaluation structure of MUTATE.} MUTATE separates action-level divergence, measured from Thought-Action attempts, from problem-level divergence, measured by distinct solution paths discovered for the same goal.}
\vspace{-4mm}
\label{fig:mutate}
\end{figure*}

\subsection{Agent Evaluation in Interactive Environments}
A separate line of work evaluates LLMs as agents in interactive environments, spanning text-based sandboxes~\cite{cote2018textworld}, web agents~\cite{zhou2024webarena, NEURIPS2022_82ad13ec, xie2024osworldbenchmarkingmultimodalagents}, embodied agents~\cite{yang2025embodiedbench, zheng2024towards}, and physical tool understanding~\cite{zhang2025phystoolbench}. Because their success conditions center on goal completion, creative and divergent problem-solving largely falls outside their scope.
Recent work has begun to introduce creativity into this setting: \citet{tian2024macgyver} constructed everyday problem scenarios requiring unconventional tool use, \citet{qian2024escapebench} extended this to interactive escape-room games with long-horizon action sequences, and multi-agent systems have explored divergent exploration, iterative refinement, and collaborative synthesis to promote creative outputs~\cite{lu2024llm, goes2023pushing, lin2025creativity, 10.1145/3715928.3737479, park2026marchevaluatingintersectionambiguity}. However, these works either do not evaluate the divergence of the step-level Thought-Action trajectory an agent produces while interacting with the environment, or do not treat the diversity of multiple solution paths discoverable for the same goal as an object of measurement. We bridge this gap with a benchmark that simultaneously formalizes and quantifies both step-level behavioral and path-level diversity within a unified framework.

\section{MUTATE}
We propose \textbf{MUTATE} (\textbf{M}ulti-path \textbf{U}nconventional \textbf{T}ask \textbf{A}ssessment with \textbf{T}ool-\textbf{E}mbedded reasoning), an interactive benchmark designed to measure divergent thinking in LLM agents.
As conceptualized in Figure~\ref{fig:mutate}, unlike conventional environments evaluating linear execution toward a single predefined solution, MUTATE places agents in text-based sandboxes where standard, intuitive approaches deliberately lead to dead-ends, forcing the agent to exhibit both action-level and path-level divergence to discover unconventional alternative paths.
\setlist{itemsep=3pt, topsep=3pt, partopsep=0pt, parsep=0pt, leftmargin=*}
\subsection{Task Formulation and Metrics}
\label{sec:task}
\paragraph{Two Levels of Agentic Divergence.}
We formalize agentic divergent thinking into two levels, designing one distinct metric for each.
\textit{Problem-level divergence} assesses whether an agent can resiliently branch away from unexpected dead-ends to discover multiple, structurally distinct solution paths toward a single fixed goal.
\textit{Action-level divergence} evaluates whether the agent's individual Thought-Action steps, including failed attempts, break conventional object affordances through mechanism-shifting ideation.
The two levels are not redundant: action-level breadth is a necessary but not sufficient condition for problem-level discovery, as a non-obvious action does not automatically constitute a goal-recoverable path.

\paragraph{Metric 1: Solution Path Discovery.}
This metric counts, across multiple trials of the same scenario, how many of the predefined valid solution paths the agent discovers.
Each scenario defines four mechanism-distinct paths per phase (\S\ref{sec:scenario}), so the metric directly captures problem-level divergence rather than mere repetition.

\paragraph{Metric 2: Divergence Momentum.}
This metric evaluates action-level divergence over \texttt{apply}, \texttt{craft}, and \texttt{input} attempts that did not lead to immediate success.
The clearest window into an agent's divergent reasoning is the \textit{moment} of action under environmental resistance, and even attempts that do not complete a path can reflect strong divergent thinking.
We implement Metric~2 via an LLM-as-a-judge along three criteria:
\textit{\textbf{Originality}} (departure from the conventional use of the target), \textit{\textbf{Elaboration}} (mechanistic depth of the Thought), and \textit{\textbf{Groundedness}} (whether the Thought justifies the necessity of the Action).
Originality and Elaboration follow established psychometric traditions~\cite{guilford1967creativity, kim2006can}, while Groundedness captures the operational validity of the agentic loop. 
The two metrics operate jointly: completed solution paths flow into Metric~1, and every creative attempt off the solution path flows into Metric~2, leaving no part of the trajectory outside the scope of evaluation.
Details including Human-LLM correlation are provided in Appendix~\ref{app:eval-reliability}.

\subsection{Interactive Environment Design}
\label{sec:scenario}

\paragraph{Simulation Engine.}
MUTATE is built as a text-based, long-horizon interactive environment in the spirit of escape-room simulations: the agent navigates between scenes, manipulates items, and crafts tools to reach a fixed scenario objective, without any external hint about which mechanism the goal requires. Concretely, we build on the interactive simulation engine of~\citet{qian2024escapebench}, consisting of three components: (1) \textit{Scenes}, unit of locations connected by movement edges; (2) \textit{Items}, interactable targets with internal states; and (3) \textit{Tools}, collectible objects stored in the agent's \textit{bag} that can be applied to items or crafted together.
At each step, the agent observes the environment and issues one of five actions: \texttt{move(scene)}, \texttt{click(item\_or\_tool)}, \texttt{apply(tool,item)}, \texttt{craft(base\_tool, ingredient\_tool)}, or \texttt{input(string,item)}. The \texttt{craft} action advances the state of \texttt{base\_tool} and is order-sensitive.
\paragraph{Scenarios.} MUTATE consists of ten custom, manually authored, and cross-validated scenarios spanning a deliberately broad set of physical settings to prevent solutions from converging on a single domain prior. Each scenario is organized as either a single phase or two phases separated by a checkpoint, and every phase contains four mechanism-distinct solution paths classified into three categories: Affordance-based (A, directly exploiting a tool's physical property), Alternative-principle (B, achieving the same goal through a different physical or chemical mechanism), and Creative-leap (C, indirect routes such as information inference or environmental manipulation). Each phase contains one A path, one B path, and two C paths, so that problem-level divergence (Metric 1) is measured as the diversity of solution mechanisms rather than surface-level tool substitution within the same mechanism. All scene, item, and tool descriptions, as well as environmental feedback, convey only physical states and constraints and never expose the intended mechanism; otherwise, action-level divergence would collapse into hint exploitation rather than reflect genuine divergent reasoning.
\begin{table*}[t]
\centering
\small
\setlength{\tabcolsep}{5pt}
\renewcommand{\arraystretch}{1.1}
\begin{tabular}{l cc ccc}
\toprule
& \multicolumn{2}{c}{\textbf{Metric 1: Path Discovery}} & \multicolumn{3}{c}{\textbf{Metric 2: Divergence Momentum}} \\
\cmidrule(lr){2-3} \cmidrule(lr){4-6}
\textbf{Model} & \textbf{Overall} & \textbf{Avg Step} & \textbf{Originality} & \textbf{Elaboration} & \textbf{Groundedness} \\
\midrule
claude-sonnet-4.6              & 30 / 56 (53.6\%)          & 26.9          & \textbf{2.764} & 3.079          & 3.350 \\
claude-haiku-4.5               & 25 / 56 (44.6\%)          & \textbf{24.6} & 2.534          & \textbf{3.337} & 3.320 \\
gpt-5.4                        & 27 / 56 (48.2\%)          & 26.3          & 2.647          & 2.827          & \textbf{3.588} \\
gpt-5.4-mini                   & 9 / 56 (16.1\%)           & 29.7          & 2.461          & 2.697          & 3.446 \\
\midrule
qwen3-235b-a22b-2507           & 22 / 56 (39.3\%)          & 33.8          & 2.755          & 3.115          & 3.388 \\
qwen3-30b-a3b-instruct-2507    & 7 / 56 (12.5\%)           & 25.0          & 2.595          & 2.799          & 3.146 \\
llama-4-maverick               & 13 / 56 (23.2\%)          & 36.2          & 2.256          & 2.667          & 3.128 \\
llama-4-scout                  & 7 / 56 (12.5\%)           & 36.3          & 2.278          & 2.470          & 2.924 \\
\midrule
Avg.\ human                    & 44.75 / 56 (79.9\%)       & 34.27         & ---            & ---            & --- \\
\bottomrule
\end{tabular}
\vspace{-2mm}
\caption{
Base-agent results across eight models on MUTATE. Closed-source models occupy the upper rows, open-weights models the lower rows, and the human baseline is reported as reference. Metric~2 is not computed for the human baseline. More details in Appendix. \ref{app:metric1-human}
}
\vspace{-4mm}
\label{tab:main_results}
\end{table*}

\subsection{Baseline Evaluation Setup}
\label{sec:setup}
\paragraph{Evaluation Protocol.}
We evaluate the ten MUTATE scenarios by allowing the agent four independent attempts per scenario.
To force path-level divergence, any solution path successfully utilized in a previous attempt is blocked and treated as unavailable in subsequent trials.
A single trial terminates when the agent reaches the goal, exceeds the maximum step budget, or repeats meaningless actions for 20 consecutive steps. The length of the agent's working memory is restricted to 10 steps.

\paragraph{Models.}
For baseline calibration (\S\ref{sec:calibration}), we evaluate eight models spanning four families, each with a stronger and a weaker variant: Claude Sonnet 4.6~\cite{anthropic2026sonnet46} and Claude Haiku 4.5~\cite{anthropic2025haiku45}, GPT-5.4 and GPT-5.4-mini~\cite{singh2025openai}, Qwen3-235B-A22B-2507 and Qwen3-30B-A3B-Instruct-2507~\cite{yang2025qwen3}, and Llama-4-Maverick and Llama-4-Scout~\cite{meta2025llama4}. The base agent follows the ReAct~\cite{yao2023reactsynergizingreasoningacting} prompting style, producing a \texttt{Thought} and a single executable \texttt{Action} per step. All models are decoded greedily. We demonstrate the robustness of our findings through temperature setting. See Appendix \ref{app:robustness} for details.

\subsection{Baseline Calibration and Failure Patterns}
\label{sec:calibration}

Using the setup of \S\ref{sec:setup}, we evaluate eight frontier LLMs under the base agent to verify that MUTATE measures what it claims to and to surface the structural difficulties it imposes.

\paragraph{Baseline Results.}
Table~\ref{tab:main_results} reports per-model performance. Closed-source models lead in Path Discovery, with Claude Sonnet 4.6 and GPT-5.4 at the top, but even the strongest model remains well below the human average (53.6\% vs.\ 79.9\%). Crucially, the two metrics do not co-vary: Claude leads Path Discovery and Originality, while top scores on Elaboration and Groundedness belong to different models. This dispersion empirically confirms the framing of \S\ref{sec:task}---action-level breadth is necessary but not sufficient for problem-level discovery.

\paragraph{Two recurring reasoning failures.}
Beneath the aggregate numbers, two failure patterns recur across every model regardless of family or scale. The first is \textit{mechanism shift failure}: when one problem-solving principle is blocked, models substitute tools within the same mechanism rather than transition to a qualitatively different one. The second is \textit{cascade construction failure}: when the correct chain requires the tool's point of application to lie at an intermediate environmental state rather than directly on the goal, models construct the necessary tool but then apply it to the most visible target, collapsing the detour. Both failures sit at the conversion point between Divergence Momentum and Path Discovery---the moment where action-level ideation must be redirected or extended into a mechanism-distinct path. ReAct-style history, which only stacks past flat triple steps, gives the agent no structured signal about \textit{which} target is accumulating failures, so the same mechanism is retried after repeated environmental resistance.

\section{ReDNA}
\label{sec:method}
\begin{figure*}[!t]
\centering
\includegraphics[width=\textwidth]{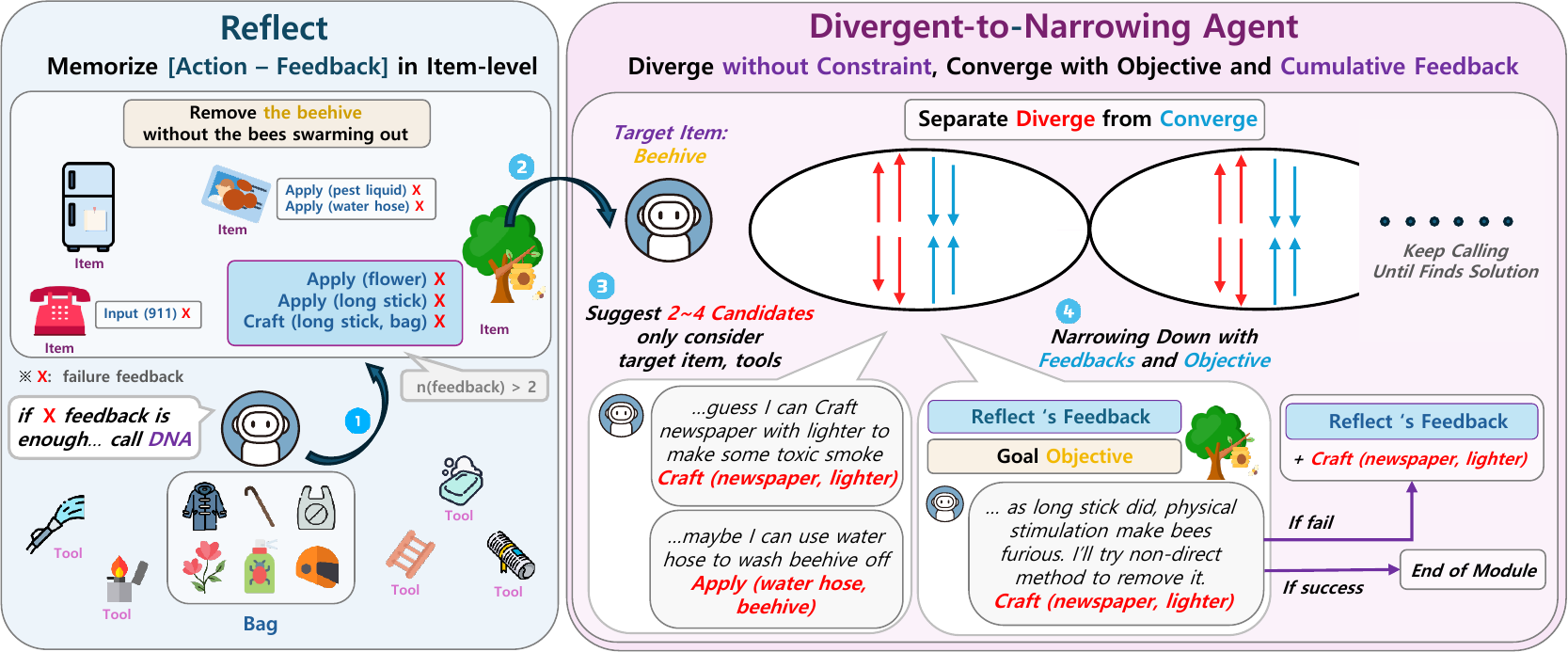}
\caption{
Architecture of ReDNA. Reflect accumulates object-level failure feedback. DN module generates candidates without convergence pressure and then selects an action based on the goal and accumulated feedback.
}
\vspace{-3mm}
\label{fig:method}
\end{figure*}

The diagnosis in \S\ref{sec:calibration} identifies a common origin for the two recurring failures: divergent ideation and convergent selection operate under the same input conditioning, so divergence is never free of the pressure to converge on the objective. We propose \textbf{ReDNA} (\textit{Reflect-driven Divergent-to-Narrowing Agent}), which enforces \textit{structural insulation} between the two operations by separating the inputs each operation is conditioned on. As shown in Figure~\ref{fig:method}, ReDNA consists of two components. The \textit{Reflect} module re-organizes the trajectory into an object-indexed failure memory, and the \textit{Diverge-then-Narrowing} (DN) module performs divergence and convergence as two phases whose inputs are deliberately separated.

\subsection{Reflect Module}

Reflect converts the raw chronological trajectory into a \textit{target-centered} failure memory. Each entry is keyed by the target item, storing the failed action, environment response, and a repetition count incremented when the identical action--response pair recurs rather than being appended as a duplicate. The memory is updated after every step in which the agent's action targets a scene item---namely \texttt{click}, \texttt{apply}, or \texttt{input}---and the environment indicates failure. The illustration in Figure~\ref{fig:method} (left) shows this organization: under the \textit{beehive} target, the agent has accumulated failures from \texttt{apply(pest liquid)}, \texttt{input(911)}, and \texttt{craft(long stick, plastic bag)}, each grouped under the same target rather than scattered across history.

The DN module is invoked when the count of \emph{substantive} failures on a single target---those produced by \texttt{apply}, \texttt{craft}, or \texttt{input} exceeds a threshold (see Appendix~\ref{app:redna}). The point at which divergent reasoning intervenes is therefore not pre-specified by a human, but is determined directly from where the agent's own trajectory has stalled.


\subsection{Divergent-to-Narrowing Module}

DN module temporarily replaces the policy and performs divergence and convergence as two phases.

\paragraph{Diverge Phase.}
This phase receives only the contents of the current scene, tools in the bag, and available actions. The scenario goal and the accumulated failure feedback are intentionally withheld. The model asks only ``what is possible with these tools in this environment'' and generates 2--4 candidate (Thought, Action) pairs. Because neither the goal nor failure feedback is present, the model avoids premature convergence---in which candidates are filtered out through self-censorship before being proposed---and can freely explore the affordance space. In Figure~\ref{fig:method} (center), the Diverge phase proposes mechanisms such as \texttt{craft(newspaper, lighter)} to generate smoke and \texttt{apply(water hose, beehive)} to wash the hive off.

\paragraph{Narrowing Phase.}
This phase reintroduces the constraints that were withheld in Diverge: the candidate (Thought, Action) pairs are reunited with the scenario goal and the target-centered failure feedback that Reflect has accumulated on the current stalled item. These constraints are not external priors but the result of the agent's own interactions with the environment. The model uses them as anchors to select one of the divergent candidates. In Figure~\ref{fig:method} (right), the water-hose candidate is filtered out---physical stimulation makes the bees more furious, consistent with the accumulated failures---and the smoke-based candidate is selected as a non-direct mechanism for the same target.

\section{Experiments}

\subsection{Main Results}
\label{sec:main_results}
\begin{table*}[t]
\centering
\small
\setlength{\tabcolsep}{5pt}
\renewcommand{\arraystretch}{1.15}
\begin{tabular*}{\textwidth}{@{\extracolsep{\fill}} l l l l l l l}
\toprule
& &
\multicolumn{2}{c}{\textbf{Metric 1: Path Discovery}}
& \multicolumn{3}{c}{\textbf{Metric 2: Divergence Momentum}} \\
\cmidrule(lr){3-4} \cmidrule(lr){5-7}
\textbf{Model} & \textbf{Method}
  & \textbf{Overall}
  & \textbf{Avg Step}
  & \textbf{Originality}
  & \textbf{Elaboration}
  & \textbf{Groundedness} \\
\midrule
\multirow{4}{*}{GPT-5.4}
  & Base
    & 27 / 56 (48.2\%) \base
    & \textbf{26.3} \base
    & 2.647 \base
    & 2.827 \base
    & 3.588 \base \\
  & Self-Refine
    & 29 / 56 (51.8\%) \gup{3.6}
    & 33.0 \rup{6.7}
    & 2.738 \gup{0.09}
    & 2.703 \rdn{0.12}
    & 3.514 \rdn{0.07} \\
  & EscapeAgent
    & 35 / 56 (62.5\%) \gup{14.3}
    & 44.9 \rup{18.6}
    & 2.542 \rdn{0.11}
    & 2.990 \gup{0.16}
    & 3.516 \rdn{0.07} \\
  & ReDNA
    & \textbf{38 / 56 (67.9\%)} \gup{19.7}
    & 40.2 \rup{13.9}
    & \textbf{2.771} \gup{0.12}
    & \textbf{3.262} \gup{0.44}
    & \textbf{3.725} \gup{0.14} \\
\midrule
\multirow{4}{*}{Claude Sonnet 4.6}
  & Base
    & 30 / 56 (53.6\%) \base
    & 26.9 \base
    & \textbf{2.764} \base
    & 3.079 \base
    & 3.350 \base \\
  & Self-Refine
    & 25 / 56 (44.6\%) \rdn{9.0}
    & \textbf{21.9} \gdn{5.0}
    & 2.733 \rdn{0.03}
    & 3.088 \gup{0.01}
    & 3.457 \gup{0.11} \\
  & EscapeAgent
    & 37 / 56 (66.1\%) \gup{12.5}
    & 43.9 \rup{17.0}
    & 2.526 \rdn{0.24}
    & 2.743 \rdn{0.34}
    & 3.341 \rdn{0.01} \\
  & ReDNA
    & \textbf{41 / 56 (73.2\%)} \gup{19.6}
    & 32.8 \rup{5.9}
    & 2.616 \rdn{0.15}
    & \textbf{3.205} \gup{0.13}
    & \textbf{3.533} \gup{0.18} \\
\midrule
\multirow{4}{*}{Llama 4 Maverick}
  & Base
    & 13 / 56 (23.2\%) \base
    & 36.2 \base
    & 2.256 \base
    & 2.667 \base
    & 3.128 \base \\
  & Self-Refine
    & 13 / 56 (23.2\%) \textsubscript{\color{gray}0.00}
    & \textbf{31.2} \gdn{5.0}
    & 2.324 \gup{0.07}
    & 2.681 \gup{0.01}
    & \textbf{3.181} \gup{0.05} \\
  & EscapeAgent
    & 23 / 56 (41.1\%) \gup{17.9}
    & 52.7 \rup{16.5}
    & 2.214 \rdn{0.04}
    & 2.571 \rdn{0.10}
    & 3.110 \rdn{0.02} \\
  & ReDNA
    & \textbf{29 / 56 (51.8\%)} \gup{28.6}
    & 45.3 \rup{9.1}
    & \textbf{2.409} \gup{0.15}
    & \textbf{3.021} \gup{0.35}
    & 3.031 \rdn{0.10} \\
\midrule
\multirow{4}{*}{Qwen3 235B A22B}
  & Base
    & 22 / 56 (39.3\%) \base
    & \textbf{33.8} \base
    & \textbf{2.755} \base
    & 3.115 \base
    & 3.388 \base \\
  & Self-Refine
    & 27 / 56 (48.2\%) \gup{8.9}
    & 35.7 \rup{1.9}
    & 2.724 \rdn{0.03}
    & 3.181 \gup{0.07}
    & 3.449 \gup{0.06} \\
  & EscapeAgent
    & 34 / 56 (60.7\%) \gup{21.4}
    & 48.3 \rup{14.5}
    & 2.645 \rdn{0.11}
    & 3.148 \gup{0.03}
    & \textbf{3.470} \gup{0.08} \\
  & ReDNA
    & \textbf{36 / 56 (64.3\%)} \gup{25.0}
    & 36.5 \rup{2.7}
    & 2.707 \rdn{0.05}
    & \textbf{3.407} \gup{0.29}
    & 3.353 \rdn{0.04} \\
\bottomrule
\end{tabular*}
\vspace{-1.5mm}
\caption{
Method comparison across models. Subscripts show the change relative to the same model's Base; color indicates whether the change is an improvement (\textcolor{improve}{green}) or degradation (\textcolor{degrade}{red}), accounting for each metric's preferred direction. Bold marks the best score within each model block.
}
\vspace{-4mm}
\label{tab:method_comparison}
\end{table*}

Table~\ref{tab:method_comparison} compares Base, Self-Refine (Appendix \ref{app:selfrefine}), EscapeAgent (Appendix \ref{app:escapeagent}), and ReDNA across the four stronger-variant models. The interpretive frame follows from \S\ref{sec:calibration}: because action-level breadth and goal-aware path discovery are distinct facets of divergence, a method that improves divergent thinking can play one of two roles---refining an already-broad ideation toward mechanism shifts, or expanding a narrow ideation space so that mechanism-distinct candidates exist in the first place. ReDNA's two phases are constructed for these two operations.

\paragraph{ReDNA improves Path Discovery across all models.}
ReDNA improves Path Discovery on all four models. The improvement holds across closed-source and open-weight models alike, indicating that the effect is not tied to a particular family or to frontier-only capability.

\paragraph{Prior methods broaden search but do not deepen divergence.}
EscapeAgent also improves Path Discovery, but its signature differs: Originality drops in every model, and Avg Step rises sharply. The added coverage comes from broader environmental exploration---more actions attempted within already-considered mechanisms---rather than from reaching qualitatively new ones. This is the mechanism shift failure of \S\ref{sec:calibration} scaled up by search rather than resolved. Self-Refine is inconsistent across models and degrades Claude Sonnet's Path Discovery, showing that a generic feasibility check is insufficient without explicit divergent--convergent structure. These results show that broadening search and deepening action-level divergence are distinct dimensions of agent behavior.

\paragraph{Elaboration improves universally under ReDNA.}
The cleanest ReDNA-specific signal appears in Elaboration: ReDNA improves it on all four models, while EscapeAgent and Self-Refine produce inconsistent or negative effects, including a substantial Elaboration drop on Claude Sonnet under EscapeAgent. ReDNA does more than expand exploration; it produces reasoning of greater mechanistic depth at each action step.


\begin{figure*}[t]
\centering
\footnotesize
\setlength{\tabcolsep}{6pt}
\renewcommand{\arraystretch}{1.15}

\begin{tabular}{
>{\raggedright\arraybackslash}p{0.485\textwidth}
@{\hspace{6pt}}
>{\raggedright\arraybackslash}p{0.485\textwidth}
}
\toprule

\rowcolor{black!8}
\multicolumn{2}{c}{%
\begin{minipage}{0.94\textwidth}
\centering
\textbf{Scenario: Zombie Lab.} Remove the infected crowd blocking the lab entrance. 
Mechanism-distinct paths: A~sound lure, B~smell disguise, C1~vent bypass, C2~incendiary dispersion.
\end{minipage}
} \\

\midrule

\rowcolor{blue!8}
\multicolumn{1}{c}{\textbf{Claude Sonnet 4.6} \textit{(high-ideation)}} &
\multicolumn{1}{c}{\textbf{GPT-5.4} \textit{(narrow-ideation)}} \\

\midrule

\rowcolor{gray!8}
\textbf{Base.} Inspects many objects (gown, rag, mask, bottle, solvent, cloth), but remains within single tool-on-target actions and never completes a construction. &
\textbf{Base.} Anchors on direct interactions with the crowd: extinguisher, solvent, and pistol are all applied directly. It never assembles a delivery tool. \\

\rowcolor{orange!12}
\multicolumn{2}{c}{\textbf{ReDNA invoked}} \\

\rowcolor{yellow!10}
\textbf{Reflect.} \texttt{apply(rag, crowd)}: odor is insufficient. \texttt{apply(extinguisher, crowd)}: the crowd scatters only temporarily. &
\textbf{Reflect.} \texttt{apply(extinguisher, crowd)}: scattering is brief and not sustained. \textit{Journal cue:} ``fire cleared the door.'' \\

\rowcolor{cyan!8}
\textbf{Diverge.}
$\bullet$ enhance the bottle/fire route\quad
$\bullet$ use solvent as a flammable source\quad
$\bullet$ craft gown + rag as a wearable disguise &
\textbf{Diverge.}
$\bullet$ repeat direct fire-tool use\quad
$\bullet$ use solvent directly\quad
$\bullet$ construct a fire-delivery tool from bottle + solvent + cloth \\

\rowcolor{green!8}
\textbf{Narrowing.} The rag-only attempt failed due to insufficient \textit{coverage}, not because the mechanism itself was wrong. The gown can provide coverage, while the rag provides odor. &
\textbf{Narrowing.} Direct extinguisher use failed. The cue implies that fire must be \textit{delivered}, not merely \textit{applied}. \\

\rowcolor{red!8}
\textbf{Action.}\newline
\texttt{craft(lab gown, infested waste rag)}\newline
\texttt{craft(lab gown, respirator mask)}\newline 
\texttt{apply(lab gown, crowd)}\newline
$\Rightarrow$ \textit{New path: B (smell disguise)} &
\textbf{Action.}\newline
\texttt{craft(glass bottle, solvent)}\newline
\texttt{craft(glass bottle, cloth strip)}\newline
\texttt{apply(glass bottle, crowd)}\newline
$\Rightarrow$ \textit{New path: C2 (incendiary)} \\

\rowcolor{black!5}
\textbf{Interpretation.} Narrowing-dominant: broad ideation is reorganized into a completed construction. &
\textbf{Interpretation.} Diverge-dominant: a construction candidate absent from the base candidate space is newly introduced. \\

\bottomrule
\end{tabular}
\vspace{-1mm}
\caption{
Case study of ReDNA on \textit{Zombie Lab}. Claude benefits from Narrowing, while GPT benefits from Diverge.
}
\vspace{-3mm}
\label{fig:case_study}
\end{figure*}
\paragraph{Two ideation profiles.}
Models split into two profiles by their base behavior (per-family analyses in Appendix~\ref{app:per-model}): a \textit{high-ideation} group (Claude Sonnet 4.6, Qwen3-235B) that produces broad candidates but rotates within the same mechanism, and a \textit{narrow-ideation} group (GPT-5.4, Llama-4-Maverick) that anchors on immediate subgoals. The two profiles differ in which side of the divergence ladder is bottlenecked.

\paragraph{ReDNA acts through different phases on each profile.}
For high-ideation models, the Narrowing phase carries weight: among the candidates the base agent already generates within a single mechanism, those exhausted by accumulated failure are excluded, and selection is redirected toward mechanism-distinct alternatives the agent's own ideation already contains but does not deploy. The Originality decrease in this group reflects this redirection of breadth into goal-aware mechanism shifts. For narrow-ideation models, the Diverge phase carries weight: candidates produced without goal or failure pressure introduce actions the base agent would not propose under standard conditioning, and Narrowing then binds them to the goal.

\paragraph{Case study.} Figure~\ref{fig:case_study} traces both ideation profiles on Zombie Lab. The two models fail in opposite ways at Base. Claude, in its high-ideation mode, inspects the gown, rag, mask, bottle, and solvent but never assembles them, so component candidates remain in the Thought line without becoming a construction. GPT, in its narrow mode, anchors on direct interactions with the crowd---extinguisher, solvent, pistol---and never proposes any delivery tool. ReDNA acts on each through a different phase. 

For Claude, Reflect accumulates the rag and extinguisher failures, and Narrowing reframes them: the rag-only attempt failed for insufficient odor coverage, not because smell disguise was the wrong mechanism, so the gown is recruited to provide the missing coverage and the path completes as smell disguise (B). For GPT, the journal cue ``fire cleared the door'' is present in the trajectory but the Base agent reads it as license to apply the extinguisher directly; Diverge, freed from goal pressure, surfaces the missing candidate of constructing a fire-delivery tool from bottle, solvent, and cloth, and Narrowing binds it to the cue and the prior extinguisher failure, completing incendiary dispersion (C2). The two models therefore reach different mechanism-distinct paths on the same scenario, illustrating that ReDNA does not push agents toward a canonical solution but unblocks the side of the divergence ladder each profile leaves unused.

\subsection{Ablation Study}
\label{sec:ablation}

ReDNA combines two components: Reflect, which accumulates object-level failure feedback, and DNA, which separates divergent candidate generation from convergent selection. To identify the source of the gains, we compare Base, an R-only ablation that retains Reflect but disables DNA, and the full ReDNA on GPT-5.4 (Table~\ref{tab:ablation}).

\begin{table}[h]
\centering
\small
\setlength{\tabcolsep}{3pt}
\renewcommand{\arraystretch}{1.2}
\begin{tabular}{lccccc}
\toprule
\textbf{Condition} & \textbf{Overall} & \textbf{Step} & \textbf{Orig.} & \textbf{Elab.} & \textbf{Ground.} \\
\midrule
Base   & 27 / 56 (48.2\%)          & 25.3 & 2.67          & 2.73          & 3.28 \\
Re only & 33 / 56 (58.9\%)          & 39.2 & 2.70          & 2.73          & 3.52 \\
ReDNA  & \textbf{38 / 56 (67.9\%)} & 36.0 & \textbf{2.77} & \textbf{3.26} & \textbf{3.72} \\
\bottomrule
\end{tabular}
\vspace{-1.5mm}
\caption{
Ablation of ReDNA components on GPT-5.4. \emph{R only} retains the Reflect module but disables the divergent-convergent (DNA) module.
}
\vspace{-4mm}
\label{tab:ablation}
\end{table}

\paragraph{Reflect alone expands opportunity, not quality.}
Re-only improves Path Discovery over Base, showing that reducing repeated failures on the same object opens opportunities to reach new paths. However, action-level metrics change only marginally. Reflect, therefore, acts as an exploration enabler: it removes wasted attempts but does not change how the agent reasons within each attempt.

\paragraph{DNA recovers divergent ideation into target paths.}
Adding DNA on top of Reflect further improves Path Discovery and produces the action-level quality gains that Reflect alone does not deliver. The mechanism is the input separation of \S\ref{sec:method}. Action-level ideation broadens, yet the action ultimately taken is recovered toward the target path. The contribution of ReDNA is therefore not failure memory itself---which Reflect alone already provides---but the structural separation that lets divergence operate without convergence pressure before convergence reshapes the result.

\subsection{Transfer to the Alternative Uses Test} \label{sec:aut}
To test whether ReDNA's input-separation principle generalizes beyond interactive settings, we apply it to the Alternative Uses Test (AUT)---a single-turn psychometric task that asks for unconventional uses of a common object~\citep{guilford1967creativity}.
We map the principle directly: the Diverge call conditions only on the object, and the Narrowing call reintroduces the creativity criterion alongside previously generated uses, which act as the non-interactive analog of failure memory.
As shown in Table~\ref{tab:aut_transfer}, this yields $+0.86$ Originality and $+0.61$ Elaboration, suggesting that the divergent--convergent decoupling, not MUTATE-specific feedback, drives the gains. Details are in Appendix~\ref{app:aut-details}
\begin{table}[t]
\centering
\small
\renewcommand{\arraystretch}{1.1}
\setlength{\tabcolsep}{12pt}
\begin{tabular}{lccc}
\toprule
\textbf{Metric} & \textbf{Base} & \textbf{ReDNA} & \textbf{Δ} \\
\midrule
Originality  & 2.91 & \textbf{3.77} & +0.86 \\
Elaboration  & 4.36 & \textbf{4.97} & +0.61 \\
\bottomrule
\end{tabular}
\vspace{-1.5mm}
\caption{
Transfer to the AUT. ReDNA improves quality-oriented dimensions (Originality, Elaboration).
}
\vspace{-4mm}
\label{tab:aut_transfer}
\end{table}

\section{Conclusion}
We introduce MUTATE, a benchmark for evaluating agentic divergent thinking via multi-path goal discovery and step-level Thought-Action divergence. MUTATE reveals two recurring failures across eight frontier models, both suggesting that agents conflate divergent generation with convergent selection. We therefore propose ReDNA, which separates candidate generation from action selection with object-level failure memory. 
ReDNA improves both divergence levels, generalizes to the Alternative Uses Test, and ablations show that structural separation drives the gains. 

\section*{Limitations}
The benchmark is based on ten human-designed scenarios and predefined gold paths, and thus needs to be validated in more diverse environments and larger-scale task sets. In addition, because step-level evaluation relies on LLM-as-a-judge, future work should strengthen evaluation reliability through larger-scale human evaluation and judge calibration. Nevertheless, this work shows that divergent thinking in LLM agents can be measured not only by success rate, but also through the joint lens of multiple path discovery and action-level ideation, and suggests that feedback-guided divergent-convergent reasoning is a promising direction for improving agentic creativity.

\section*{Ethics Statement}
This work uses manually authored, simulated text environments to study divergent thinking in LLM agents. MUTATE contains no private user data or personally identifying information, and all agent interactions occur within a constrained sandbox of predefined scenes, objects, actions, and solution paths. We evaluate open-weight and API-based LLMs under documented access settings and follow the providers' usage terms. For reliability analysis, we conduct human studies in which participants solve synthetic benchmark scenarios and annotators score sampled Thought--Action records; annotators are compensated, and the results are reported only in aggregate. Because some scenarios involve fictional physical problem-solving situations, the benchmark and ReDNA are intended solely for research evaluation, not for real-world repair, emergency, or safety guidance. We also acknowledge possible subjectivity in LLM-as-a-judge scoring and therefore validate it against human annotations.





\section*{Acknowledgement}
This work was supported by the Institute of Information \& Communications Technology Planning \& Evaluation (IITP) grant funded by the Korea government (MSIT) [RS-2021-II211341, Artificial Intelligence Graduate School Program (Chung-Ang University)] and the National Research Foundation of Korea(NRF) grant funded by the Korea government(MSIT) (RS-2025-24683575). This work was also supported by the National Research Foundation of Korea (NRF) grant funded by the Korean government (MSIT) (RS-2026-25494299).
\bibliography{custom}
\clearpage
\appendix
\lstdefinestyle{appendixprompt}{
    basicstyle=\footnotesize\ttfamily,
    breaklines=true,
    breakatwhitespace=false,
    breakindent=1.2em,
    columns=fullflexible,
    keepspaces=true,
    frame=none,
    xleftmargin=0.9em,
    xrightmargin=0.9em,
    aboveskip=0.35em,
    belowskip=1.0em,
    showstringspaces=false
  }
\newcommand{\okmark}{O}
\newcommand{\xmarktab}{X}

\section{Appendix Table of Contents}
\label{app:toc}
We organize the Appendix as follows:

\begin{enumerate}[leftmargin=*, itemsep=2pt, topsep=2pt]
    \item Section~\ref{app:per-model} provides per-model reasoning style analyses, including representative trajectory excerpts for GPT, Claude, Llama, and Qwen families.

    \item Section~\ref{app:eval-reliability} validates the evaluation metrics, including LLM-as-a-judge agreement with human annotations and the human solving protocol for Metric~1.

    \item Section~\ref{app:method-details} provides implementation details for the compared methods, including the Base agent, Self-Refine, EscapeAgent, and ReDNA.

    \item Section~\ref{app:mutate-details} describes the MUTATE benchmark, including the YAML scenario structure, solution-path catalog, per-scenario path discovery, and the human-evaluation UI.

    \item Section~\ref{appn:prompt} lists the full prompt templates used for the Base agent, EscapeAgent, Self-Refine, and ReDNA.
\end{enumerate}

\section{Per-Model Reasoning Style Analysis}
\label{app:per-model}

Section~\ref{sec:main_results} groups the eight models into two action-level ideation profiles—high-ideation (Claude Sonnet 4.6, Qwen3-235B) and narrow-ideation (GPT-5.4, Llama-4-Maverick)—and shows that ReDNA acts through different phases on each profile. Here we provide finer-grained reasoning style descriptions within each profile, supported by representative trajectory excerpts drawn from the base-agent runs. Table~\ref{tab:model-reasoning-summary} summarizes the profiles; \S\ref{app:per-model-gpt}--\S\ref{app:per-model-qwen} discuss each family with excerpts.

\begin{table*}[t]
\centering
\small
\setlength{\tabcolsep}{4pt}
\renewcommand{\arraystretch}{1.25}
\begin{tabularx}{\textwidth}{p{0.12\textwidth} p{0.22\textwidth} X X}
\toprule
\textbf{Family} & \textbf{Style} & \textbf{Strengths} & \textbf{Limitations} \\
\midrule
GPT & Goal-aligned and conservative & Goal alignment; action grammar; physical repair sequences; low random tool use; stable groundedness. & Narrow mechanism space; delayed pivot; repeated local variants; weak multi-object construction; limited indirect delegation or symbolic transformation. \\
\midrule
Claude & High-ideation and verbally rich & Rich elaboration; fluent candidate generation; strong affordance language; many possible tool uses; high originality. & Weak pruning; tool substitution within one target; soft use of failure feedback; delayed mechanism pivot; unstable stateful construction. \\
\midrule
Llama & Exploratory but unstable & Active scene exploration; willingness to try alternatives; broad object probing; fast response on simple affordances; benefits from scaffolding. & Unstable feedback integration; weak Thought--Action alignment; skipped construction steps; loose objective connection; limited mechanism-level filtering. \\
\midrule
Qwen & Mechanism-inventive but weakly disciplined & High originality; non-obvious affordance reinterpretation; mechanism construction; broad candidate space; strong benefit from failure grounding. & Weak convergence discipline; invalid tool-state assumptions; loose grammar constraints; over-wide mechanisms; unstable coverage despite originality. \\
\bottomrule
\end{tabularx}
\caption{Reasoning profiles across the four model families. The high-ideation profile (Claude, Qwen) is bottlenecked by convergence to mechanism-distinct paths; the narrow-ideation profile (GPT, Llama) is bottlenecked by candidate breadth.}
\label{tab:model-reasoning-summary}
\end{table*}

\subsection{GPT: Goal-Aligned Local Repair}
\label{app:per-model-gpt}

GPT models show goal-aligned, conservative, and locally coherent reasoning. The model identifies the current goal, selects a direct subgoal, chooses a grounded action, and, after failure, repairs the current hypothesis with a close variant rather than pivoting to a different mechanism.

Table~\ref{tab:excerpt-gpt} shows two representative excerpts. In Excerpt~1, goal alignment is high, and the model grounds its action on an explicit clue (the reversed sticky note). However, the clue--object assignment is wrong, and GPT continues to generate code variants \emph{within the same hypothesis} rather than reassigning the clue to a different target or shifting to a non-input mechanism. This is the \emph{cascade construction failure} of \S\ref{sec:calibration} in concrete form: the construction is locally coherent, but the mechanism is not revisited. In Excerpt~2, the model correctly extracts the ``fire'' clue from the journal and selects the most fire-related object available. The action is grounded, but it does not extend to constructing a \emph{fire-delivery} object (bottle + solvent + wick), which is what the C2 path actually requires. ReDNA helps GPT primarily by expanding the candidate mechanism space in the Diverge phase before convergence selects a valid action.

\begin{table}[tbp]
\centering
\footnotesize
\setlength{\tabcolsep}{4pt}
\renewcommand{\arraystretch}{1.3}
\begin{tabularx}{\columnwidth}{@{}c X p{0.32\columnwidth}@{}}
\toprule
\textbf{\#} & \textbf{Thought} & \textbf{Action} \\
\midrule
\multicolumn{3}{@{}l}{\textit{Excerpt 1 — \texttt{house\_heist / run\_3}}} \\
\midrule
1 & The reversed sticky note likely gives the safe code, so I should use that clue instead of guessing more numbers. & \texttt{input(2518,}\newline\texttt{\ \ fingerprint safe)} \\
2 & The reversed sticky note likely gives the 8-digit spare key code, so I should try that on the safe keypad. & \texttt{input(25158108,}\newline\texttt{\ \ fingerprint safe)} \\
3 & The reversed sticky note likely hides the safe code; entering a corrected 8-digit version is the strongest next step. & \texttt{input(25180000,}\newline\texttt{\ \ fingerprint safe)} \\
\midrule
\multicolumn{3}{@{}l}{\textit{Excerpt 2 — \texttt{zombie\_lab / run\_1}}} \\
\midrule
1 & The journal strongly hints fire clears the zombies, and I have a charged extinguisher to test on the crowd as the most direct way to manipulate them. & \texttt{apply(fire extinguisher,}\newline\texttt{\ \ lab entrance crowd)} \\
\bottomrule
\end{tabularx}
\caption{Representative excerpts from GPT-5.4. Excerpt~1: code-variant repetition within a single hypothesis. Excerpt~2: direct application without intermediate construction.}
\label{tab:excerpt-gpt}
\end{table}

\subsection{Claude: High-Ideation Within-Mechanism Rotation}
\label{app:per-model-claude}

Claude models produce many candidate ideas in the Thought line, with detailed affordance descriptions and plausible alternatives. The surface signal is divergent, but the chosen action often remains trapped inside the same target or mechanism family.

Table~\ref{tab:excerpt-claude} shows two representative excerpts. In Excerpt~1, the Thought line lists many candidates (detergent, metal pipe, coin, rolling basket, lucky number). However, every candidate orbits the same target (the rusty rear gate) and the same underlying mechanism (lubrication or leverage). This is the \emph{mechanism shift failure} of \S\ref{sec:calibration}: breadth exists at the ideation level but does not deploy as mechanism shifts. In Excerpt~2, object affordances are read with mechanistic precision (ridge pattern, fingerprint, adhesive tape). Even when a path is forbidden, the model searches for variants around the same construction. ReDNA helps Claude primarily by narrowing this already-broad ideation through accumulated failure constraints, redirecting breadth into mechanism-distinct deployment rather than within-mechanism rotation.

\begin{table}[tbp]
\centering
\footnotesize
\setlength{\tabcolsep}{4pt}
\renewcommand{\arraystretch}{1.3}
\begin{tabularx}{\columnwidth}{@{}c X p{0.32\columnwidth}@{}}
\toprule
\textbf{\#} & \textbf{Thought} & \textbf{Action} \\
\midrule
\multicolumn{3}{@{}l}{\textit{Excerpt 1 — \texttt{laundry\_machine\_7 / run\_4}}} \\
\midrule
1 & Rear gate is rusty and seized. Tried metal pipe, dolly, coin. Detergent is thick and slippery---could lubricate the latch, but that's forbidden. Maybe craft detergent with another tool, or try the rolling basket on the gate. & \texttt{apply(rolling basket,}\newline\texttt{\ \ rear gate)} \\
2 & Gate is rusty and seized. Detergent could lubricate the latch but forbidden. Try crafting detergent with something, or apply the coin (failed before). Let me try crafting detergent onto the metal pipe for a lubricated pipe. & \texttt{craft(metal pipe,}\newline\texttt{\ \ detergent)} \\
\midrule
\multicolumn{3}{@{}l}{\textit{Excerpt 2 — \texttt{house\_heist / run\_2}}} \\
\midrule
1 & The wine glass shows a clear ridge pattern (fingerprint). I should use adhesive tape to lift it, then use it on the safe. But \texttt{apply(adhesive tape with fingerprint, fingerprint safe)} is forbidden. Let me try crafting adhesive tape with the fingerprint from the wine glass. & \texttt{apply(adhesive tape,}\newline\texttt{\ \ wine glass)} \\
\bottomrule
\end{tabularx}
\caption{Representative excerpts from Claude Sonnet 4.6. Both excerpts show many candidates surfacing in the Thought line while the chosen Action rotates within a single mechanism family.}
\label{tab:excerpt-claude}
\end{table}

\subsection{Llama: Exploration Without Mechanism Discipline}
\label{app:per-model-llama}

Llama models actively explore scenes and objects, but the update from failure feedback to a new mechanism is less reliable. The model is not simply conservative—it explores—but its exploration is weakly structured by failure evidence, so the bottleneck appears at mechanism construction rather than at exploration breadth.

Table~\ref{tab:excerpt-llama} shows two excerpts from the same scenario. In Excerpt~1, scene exploration is active and the vending--change machine relation is correctly recognized. However, the model repeats actions on the same object and does not consolidate failure feedback into a precise mechanism update. In Excerpt~2, the Thought recognizes both the fire clue and the fact that the extinguisher is the \emph{opposite} of fire. The mechanistic insight is correct, but the action does not stabilize on the implied construction (assembling a fire-delivery object); it oscillates between re-examining the target and direct application. ReDNA helps Llama by turning scattered exploration into item-level, failure-constrained candidate selection.

\begin{table}[tbp]
\centering
\footnotesize
\setlength{\tabcolsep}{4pt}
\renewcommand{\arraystretch}{1.3}
\begin{tabularx}{\columnwidth}{@{}c X p{0.32\columnwidth}@{}}
\toprule
\textbf{\#} & \textbf{Thought} & \textbf{Action} \\
\midrule
\multicolumn{3}{@{}l}{\textit{Excerpt 1 — \texttt{zombie\_lab / run\_4}}} \\
\midrule
1 & I need to explore the current scene and progress towards retrieving the cure. The vending machine seems like a potential point of interest. & \texttt{click(vending machine)} \\
2 & I need to explore and understand how to interact with the items. The vending machine and change machine seem related, and I have a bill that could be used. & \texttt{click(vending machine)} \\
3 & The vending machine seems key to obtaining something useful, possibly by using the change machine to get coins. & \texttt{apply(bill,}\newline\texttt{\ \ change machine)} \\
\midrule
\multicolumn{3}{@{}l}{\textit{Excerpt 2 — \texttt{zombie\_lab / run\_4}}} \\
\midrule
1 & Lab entrance is blocked by zombies. The journal hinted fire was used to clear the door. I have a fire extinguisher, which is the opposite of fire, but it might create a different solution or distract the zombies. & \texttt{click(lab entrance crowd)} \\
2 & Lab entrance is blocked by zombies and a body jammed in the door. The journal hinted fire was used. I have a fire extinguisher, the opposite of fire, but it might be useful in a different way. & \texttt{apply(fire extinguisher,}\newline\texttt{\ \ lab entrance crowd)} \\
\bottomrule
\end{tabularx}
\caption{Representative excerpts from Llama-4-Maverick. Excerpt~1: action repetition despite active exploration. Excerpt~2: mechanistic insight that does not stabilize into a construction step.}
\label{tab:excerpt-llama}
\end{table}

\subsection{Qwen: Mechanism Invention Without Grammar Discipline}
\label{app:per-model-qwen}

Qwen frequently reinterprets objects as carriers, wedges, lures, insulators, signals, containers, triggers, or disguise components. The challenge is not the absence of ideas, but selecting which creative mechanism fits the current environment constraints and accumulated failure evidence.

Table~\ref{tab:excerpt-qwen} shows three excerpts spanning Qwen's characteristic patterns. In Excerpt~1, the pillow is reinterpreted as a sound dampener and the metal bookend as a mass component; the proposed mechanism is functionally coherent. Failure, however, does not redirect Qwen away from the over-wide mechanism family; it tends to extend in the same direction. In Excerpt~2, the object state is functionally reconstructed: a metal bar as a lever, grip cloth as a friction modifier. This is closer to mechanism construction than to tool substitution, and characteristic of Qwen's strength. In Excerpt~3, the odor-deception mechanism is invented by the model itself, which is characteristic of Qwen. The correct B path, however, requires a wearable disguise construction (lab gown + infected waste rag + respirator mask), not a more potent scent tool alone. Under ReDNA, Qwen's originality can be channeled toward path discovery, which can reduce measured originality in some settings because the method filters wild but weakly grounded candidates, while improving coverage by selecting task-useful mechanisms.

\begin{table}[tbp]
\centering
\footnotesize
\setlength{\tabcolsep}{4pt}
\renewcommand{\arraystretch}{1.3}
\begin{tabularx}{\columnwidth}{@{}c X p{0.32\columnwidth}@{}}
\toprule
\textbf{\#} & \textbf{Thought} & \textbf{Action} \\
\midrule
\multicolumn{3}{@{}l}{\textit{Excerpt 1 — \texttt{missed\_call / run\_1}}} \\
\midrule
1 & The music is still too loud and the small pillow alone isn't sufficient. The metal bookend is heavy and solid; combining it with the pillow might create a more effective sound-blocking tool. & \texttt{craft(small pillow,}\newline\texttt{\ \ metal bookend)} \\
\midrule
\multicolumn{3}{@{}l}{\textit{Excerpt 2 — \texttt{pipe\_leak / run\_3}}} \\
\midrule
1 & The sealed manhole needs to be opened. The metal bar could serve as a lever, but its slick surface makes it hard to grip. The grip cloth is designed for high friction; crafting it with the metal bar would create a better tool for prying. & \texttt{craft(metal bar,}\newline\texttt{\ \ grip cloth)} \\
\midrule
\multicolumn{3}{@{}l}{\textit{Excerpt 3 — \texttt{zombie\_lab / run\_2}}} \\
\midrule
1 & The solvent and infected waste rag may be crafted together to simulate a stronger odor for deception. Crafting them could create a more potent scent tool. & \texttt{craft(infected waste rag,}\newline\texttt{\ \ solvent)} \\
\bottomrule
\end{tabularx}
\caption{Representative excerpts from Qwen3-235B. Excerpts~1--2: characteristic mechanism invention. Excerpt~3: invented mechanism that does not land on the correct construction state.}
\label{tab:excerpt-qwen}
\end{table}

\paragraph{Summary across families.}
The four families differ not only in capability level but in \emph{which side of the divergence ladder is bottlenecked}. The high-ideation profile (Claude, Qwen) generates broad candidates but rotates within the same mechanism family; the narrow-ideation profile (GPT, Llama) anchors on direct hypotheses or oscillates between exploration and action without consolidating mechanism construction. The asymmetric gains under ReDNA reported in Section~\ref{sec:main_results} follow directly from this split: the Narrowing phase carries weight where ideation already exists but does not deploy, and the Diverge phase carries weight where candidates are missing in the first place.

\section{Evaluation Reliability}
\label{app:eval-reliability}

We validate the two metrics against human reference using two independent studies, one for each metric.
\subsection{Metric~1: Human Solving Protocol}
\label{app:metric1-human}

The human baseline reported in Table~\ref{tab:main_results} (44.75/56, 79.9\%) is obtained by having human participants attempt the MUTATE scenarios under the same multi-attempt protocol used for the agents (\S\ref{sec:setup}). Unlike the Metric~2 annotation, which scores existing Thought-Action records produced by models, this study has humans \textit{solve} the scenarios themselves and counts the number of distinct solution paths they discover.

\paragraph{Interface.}
Participants interact with each scenario through the benchmark UI shown in Figure~\ref{fig:eval-ui}. The interface exposes the same observation space available to the agents---current scene description, visible items, tools in the bag, valid actions (\texttt{move}, \texttt{click}, \texttt{apply}, \texttt{craft}, \texttt{input}), and environment responses---so that human and agent attempts are evaluated under matched conditions. The Login and Tutorial views (Figure~\ref{fig:eval-ui}a-b) onboard participants; the Main and Scoring views (Figure~\ref{fig:eval-ui}c-d) display the scenario state and accept the next action; and the Context and Bag views (Figure~\ref{fig:eval-ui}e-f) surface action history and current inventory.

\paragraph{Protocol.} We recruit four graduate students from our lab and neighboring research groups as participants. Each participant runs four independent attempts per scenario through the benchmark UI (Figure~\ref{fig:eval-ui}), which exposes the same observation space available to the agents---current scene description, visible items, tools in the bag, valid actions (\texttt{move}, \texttt{click}, \texttt{apply}, \texttt{craft}, \texttt{input}), and environment responses---so that human and agent attempts are evaluated under matched conditions. The multi-attempt protocol matches the agent setting: any solution path successfully utilized in a previous attempt is blocked and treated as unavailable in subsequent trials. Participants completed the full ten-scenario set in approximately six to seven hours each, and were compensated at 10 USD per hour. An attempt terminates when the participant reaches the goal, exceeds the maximum step budget, or repeats meaningless actions for 20 consecutive steps. Participants receive the scenario objective but no information about the underlying mechanism categories (A/B/C1/C2) or the existence of multiple paths beyond what the diversity constraint implies. We count, for each scenario, the number of distinct gold paths reached across the four attempts, and aggregate across scenarios to produce the 44.75/56 figure reported in the main text. We do not collect verbal Thought traces from participants due to the time and cost of per-action justification, so Metric 2 is not computed for the human baseline.

\subsection{Metric 2: LLM-as-a-Judge and Human Annotation}

Metric 2 scores non-terminal \texttt{apply}, \texttt{craft}, and \texttt{input}
attempts with an LLM-as-a-judge along the three criteria.
To verify that the judge tracks human reasoning, we collect parallel human scores on a sample of 296 Thought-Action records (yielding 888 criterion-level comparisons) and compare the two streams.

\paragraph{LLM-as-a-Judge Setup.} The judge receives the scenario objective, current scene context, tools in the bag, recent history, the model's Thought, the selected Action, and the environment response. This context is necessary because the same surface action can be creative or ungrounded depending on the available tools, the current item state, and prior feedback. We use the LLM judge for large-scale scoring, and then validate its behavior against the human annotations described below, rather than treating it as a replacement for human judgment.

\paragraph{Human Annotation Protocol.} Two graduate-level annotators fluent in English score the 296 sampled Thought-Action records under the same three criteria as the judge. Annotators are compensated at 10 USD per hour and receive written guidelines, criterion definitions, and example cases; a calibration session aligns rubric interpretation before final scoring.
Each annotation item exposes the same context shown to the judge (scenario objective, environment state, action history, bag contents, Thought, selected
Action, environment response), so that the two streams of scores are directly comparable. Annotators are asked to judge the attempt itself and its stated mechanism, not whether the action eventually solves the scenario.

\paragraph{Human–LLM Correlation Results.} Table~\ref{tab:llm-human-agreement-updated} reports Pearson correlation, Spearman rank correlation, and pairwise agreement between the LLM judge and human reference scores. On pairwise agreement, the LLM judge reaches $0.8447$ overall, which falls within the range that~\citet{chiang2023can} identify as matching ``the same level of agreement between humans'' for strong LLM judges. Per-criterion pairwise agreement is consistent with this regime, with Groundedness ($0.8988$) and
Elaboration ($0.8637$) at the higher end and Originality ($0.7717$) at the lower end. The same ordering appears in the Pearson and Spearman columns:
Groundedness and Elaboration show the strongest agreement, while Originality trails, which we interpret as expected residual subjectivity, consistent with prior observations that more subjective rubric dimensions are intrinsically harder for both humans and LLM judges to agree on~\citep{chiang2023can}.
We therefore use Metric 2 primarily for aggregate comparisons and qualitative analysis, not as a claim of perfect record-level equivalence to human judgment.

\begin{table}[t]
\centering
\footnotesize
\setlength{\tabcolsep}{3pt}
\begin{tabular}{lrrrr}
\toprule
\textbf{Criterion} & $n$ & \textbf{Pearson} & \textbf{Spearman} & \textbf{Pairwise} \\
\midrule
Overall      & 888 & 0.5564 & 0.5372 & 0.8447 \\
\midrule
Originality  & 296 & 0.4094 & 0.4025 & 0.7717 \\
Elaboration  & 296 & 0.6227 & 0.5899 & 0.8637 \\
Groundedness & 296 & 0.6371 & 0.6193 & 0.8988 \\
\bottomrule
\end{tabular}
\caption{Human--LLM agreement on the three Metric~2 criteria.}
\label{tab:llm-human-agreement-updated}
\end{table}

\section{Method Implementation Details}
\label{app:method-details}
\begin{table}[t]
\centering
\small
\renewcommand{\arraystretch}{1.15}
\setlength{\tabcolsep}{4pt}
\begin{tabular}{@{}p{0.46\textwidth}@{}}
\toprule
\textbf{Algorithm: ReDNA} \\
\midrule
\textbf{Require:} goal $g$; environment $\mathcal{E}$; base policy $\pi_{\text{base}}$; \\
\hspace{1.2em} failure threshold $\tau$; max steps $T$ \\
\textbf{Ensure:} executed trajectory $\mathcal{T}$ \\
\midrule
1:\ \ Initialize failure memory $\mathcal{M} \gets \emptyset$ \\
2:\ \ Initialize substantive counter $c[\cdot] \gets 0$ \\
3:\ \ Initialize trajectory $\mathcal{T} \gets \emptyset$ \\
4:\ \ \textbf{for} $t = 1, 2, \ldots, T$ \textbf{do} \\
5:\ \ \quad Observe scene $s_t$, bag $\mathcal{B}_t$, actions $\mathcal{A}_t$ \\
6:\ \ \quad \textbf{if} goal $g$ reached \textbf{then break} \\
\multicolumn{1}{@{}l@{}}{\textit{\quad \# Policy selection based on failures}} \\
7:\ \ \quad \textbf{if} $\exists\, I^\star$ s.t. $c[I^\star] \geq \tau$ \textbf{then} \\
8:\ \ \quad\quad \textit{\# DN module: single LLM call, two phases} \\
9:\ \ \quad\quad \textit{\# Diverge: condition on $(I^\star\!, s_t, \mathcal{B}_t, \mathcal{A}_t)$ only} \\
10:\ \ \quad\quad $\mathcal{C} \!\gets\! \textsc{Diverge}(I^\star\!, s_t, \mathcal{B}_t, \mathcal{A}_t)$ \\
11:\ \ \quad\quad \quad generate 2--4 candidates $\{(\text{th}_i, a_i)\}$ \\
12:\ \ \quad\quad \textit{\# Narrowing: reintroduce $(g, \mathcal{M}[I^\star])$} \\
13:\ \ \quad\quad $(\text{th}^\star\!, a_t) \!\gets\! \textsc{Narrow}(\mathcal{C}, g, \mathcal{M}[I^\star])$ \\
14:\ \ \quad \textbf{else} \\
15:\ \ \quad\quad $(\text{th}^\star\!, a_t) \!\gets\! \pi_{\text{base}}(s_t, \mathcal{B}_t, \mathcal{A}_t, g)$ \\
16:\ \ \quad \textbf{end if} \\
17:\ \ \quad Execute $a_t$ in $\mathcal{E}$; receive response $r_t$ \\
18:\ \ \quad Append $(\text{th}^\star\!, a_t, r_t)$ to $\mathcal{T}$ \\
\multicolumn{1}{@{}l@{}}{\textit{\quad // Reflect: target-centered memory update}} \\
19:\ \ \quad \textbf{if} $a_t$ targets item $I$ \textbf{and} $r_t$ is failure \textbf{then} \\
20:\ \ \quad\quad \textbf{if} $a_t \in \{\texttt{click},\texttt{apply},\texttt{input}\}$ \textbf{then} \\
21:\ \ \quad\quad\quad \textbf{if} $(a_t, r_t) \in \mathcal{M}[I]$ \textbf{then} \\
22:\ \ \quad\quad\quad\quad increment repetition count \\
23:\ \ \quad\quad\quad \textbf{else} $\mathcal{M}[I] \!\gets\! \mathcal{M}[I] \cup \{(a_t, r_t)\}$ \\
24:\ \ \quad\quad\quad \textbf{end if} \\
25:\ \ \quad\quad \textbf{end if} \\
26:\ \ \quad\quad \textbf{if} $a_t \in \{\texttt{apply},\texttt{craft},\texttt{input}\}$ \textbf{then} \\
27:\ \ \quad\quad\quad $c[I] \gets c[I] + 1$ \quad // substantive \\
28:\ \ \quad\quad \textbf{end if} \\
29:\ \ \quad \textbf{end if} \\
30:\ \ \textbf{end for} \\
31:\ \ \textbf{return} $\mathcal{T}$ \\
\bottomrule
\end{tabular}
\caption{ReDNA algorithm.}
\label{tab:redna_algorithm}
\end{table}

Table~\ref{tab:method-summary} summarizes the implementation differences among methods. We then describe each method in the same order.

\begin{table*}[t]
\centering
\small
\setlength{\tabcolsep}{4pt}
\renewcommand{\arraystretch}{1.08}
\begin{tabularx}{\textwidth}{p{0.14\textwidth} p{0.22\textwidth} p{0.28\textwidth} X}
\toprule
\textbf{Method} & \textbf{State} & \textbf{Intervention} & \textbf{Role} \\
\midrule
Base & Recent history only & Single action policy & Provides the standard local decision baseline. \\
Self-Refine & Recent history only & Feasibility self-check & Reduces invalid or weakly grounded actions. \\
EscapeAgent & Task list and forethought queue & Reflection plus candidate generation & Expands exploration by proposing new tool-task candidates. \\
ReDNA & Target-centered failure memory & Divergent generation followed by constrained selection & Uses failure evidence to select mechanism-level alternatives. \\
\bottomrule
\end{tabularx}
\caption{Method summary.}
\label{tab:method-summary}
\end{table*}

\subsection{Base Agent}
\label{app:baseagent}
The Base agent~\cite{yao2023reactsynergizingreasoningacting} is the standard action policy used for comparison. We implement ReAct in a zero-shot setting, where environment responses are recorded in the recent interaction history without explicit Observation labels. At each step, we provide the current scene description, possible actions, tools in the bag, recent interaction history, and the diversity constraint that prevents reuse of already discovered finish actions. The LLM outputs exactly one next action in the two-line \texttt{Thought:}/\texttt{Action}: format, following the Thought–Action structure of ReAct.

The Base agent does not maintain a task list or structured failure memory. Failed actions remain in the recent history, but we do not separately abstract which target item accumulates failures or which mechanism repeatedly fails. Its decision process is therefore local to the current scene and recent trajectory.

\subsection{Self-Refine Agent}
\label{app:selfrefine}
The Self-Refine~\cite{madaan2023self} agent adds a feasibility check to the Base prompt. We ask the model to verify, before writing the final action, whether the action is physically plausible, follows the action grammar, uses tools already in the bag, targets a current-scene item for \texttt{apply}, uses two bag tools for \texttt{craft}, uses a valid movement label for \texttt{move}, and avoids clearly repeated failures.

Self-Refine does not add external memory or new exploration state. Its role is to reduce invalid or weakly grounded actions rather than to expand the mechanism space.

\subsection{EscapeAgent}
\label{app:escapeagent}
EscapeAgent~\cite{qian2024escapebench} adds task-list reflection and forethought modules on top of the Base agent. After an action and environment response, the reflection module updates a task list with unresolved targets and feedback. When the agent collects a new tool or discovers a new task, a forethought prompt proposes possible \texttt{craft}, \texttt{apply}, \texttt{click}, or \texttt{input} candidates.

This design broadens candidate generation. However, the generated candidates are not always constrained by item-level failure evidence. EscapeAgent can therefore produce many superficially creative tool-target proposals whose functional relation to the current obstacle is weak. In \textit{Zombie Lab}, for example, EscapeAgent can discover all paths, but it also spends actions on proposals such as applying currency or protective equipment to unrelated shelves.

\subsection{ReDNA}
\label{app:redna}
ReDNA combines target-centered failure memory with a feedback-guided divergent-convergent reasoning module, formalized in Algorithm~\ref{tab:redna_algorithm}. We retain the Base policy for ordinary steps, and once the substantive failure counter $c[I^\star]$ on some target item crosses the threshold $\tau$, the DN module temporarily takes over. Table~\ref{tab:redna-procedure} summarizes the full per-step procedure, and Table~\ref{tab:redna-hyperparams} lists the hyperparameter values used in our experiments.

The reflection memory $\mathcal{M}$ stores failed interactions by the target item. For a target such as \texttt{lab entrance crowd}, failures like \texttt{apply(infected waste rag, lab entrance crowd)} and \texttt{apply(fire extinguisher, lab entrance crowd)} become accumulated evidence about the same obstacle, with identical $(a_t, r_t)$ pairs merged via a repetition count rather than appended as duplicates. The counter $c[I]$ is incremented only for \texttt{apply}, \texttt{craft}, and \texttt{input}, excluding \texttt{click} since it primarily serves observation, so DN is triggered by mechanism-level resistance rather than by inspection failures. We use this memory not only to avoid exact repetition, but also to diagnose why a mechanism fails and what missing property a new mechanism should provide.

The DN module has two phases. In the divergent phase, the model generates two to four candidate mechanisms from the current target item $I^\star$, scene, and bag tools, with the scenario goal $g$ and the failure memory $\mathcal{M}[I^\star]$ intentionally withheld to prevent premature convergence. In the convergent phase, the model reunites these candidates with the objective, possible actions, diversity constraint, and the accumulated failure memory $\mathcal{M}[I^\star]$, and selects one valid \texttt{apply}, \texttt{input}, or \texttt{craft} action. This lets ReDNA expand alternatives near the failed target while still selecting a grounded action.




\begin{table}[t]
\centering
\small
\setlength{\tabcolsep}{4pt}
\renewcommand{\arraystretch}{1.25}
\begin{tabularx}{\columnwidth}{@{}l X@{}}
\toprule
\textbf{Stage} & \textbf{Operation} \\
\midrule

\multicolumn{2}{@{}l}{\textbf{(1) Per-step memory update} (Reflect)} \\
\textit{Trigger} & every step in which the action targets a scene item \\
\textit{Record} & store $(a, r)$ under the target item if $a \in \{\texttt{click},\texttt{apply},\texttt{input}\}$ and $r$ is a failure; merge identical $(a, r)$ pairs by incrementing a count rather than appending duplicates \\
\textit{Substantive counter} & increment a separate per-target count when $a \in \{\texttt{apply},\texttt{craft},\texttt{input}\}$ (excluding \texttt{click}) \\
\midrule

\multicolumn{2}{@{}l}{\textbf{(2) DN invocation check}} \\
\textit{Condition} & DN module is invoked at the next step iff the substantive counter on some target $T$ reaches threshold $\tau$ \\
\textit{Otherwise} & the base ReAct policy decides the next action \\
\midrule

\multicolumn{2}{@{}l}{\textbf{(3) Diverge phase} (candidate generation)} \\
\textit{Inputs supplied} & target $T$; scene context; tools in bag; available actions \\
\textit{Inputs withheld} & scenario goal $g$; accumulated failure memory $\mathcal{M}[T]$ \\
\textit{Output} & 2--4 candidate (Thought, Action) pairs $\{(t_i, a_i)\}$, generated under goal- and failure-free conditioning \\
\midrule

\multicolumn{2}{@{}l}{\textbf{(4) Narrowing phase} (constraint-anchored selection)} \\
\textit{Inputs supplied} & candidate set $\{(t_i, a_i)\}$; scenario goal $g$; failure memory $\mathcal{M}[T]$ \\
\textit{Selection rule} & choose one $(t^{\star}, a^{\star})$ that is consistent with $g$ and not contradicted by entries in $\mathcal{M}[T]$ \\
\textit{Output} & a single executable action $a^{\star}$ to be issued to the environment \\

\bottomrule
\end{tabularx}
\caption{Procedure of ReDNA per environment step.}
\label{tab:redna-procedure}
\end{table}


\begin{table}[t]
\centering
\small
\setlength{\tabcolsep}{4pt}
\begin{tabularx}{\columnwidth}{p{0.42\columnwidth} X}
\toprule
\textbf{Setting} & \textbf{Value} \\
\midrule
Tracked actions & \texttt{apply}, \texttt{craft}, \texttt{input} \\
Excluded actions & \texttt{move}, \texttt{click} \\
Memory key & Target item \\
Trigger condition & Repeated same-target failure with non-click evidence \\
Candidate count & 2--4 mechanisms \\
DN final actions & \texttt{apply}, \texttt{input}, or \texttt{craft} only \\
Diversity control & Previously discovered finish actions are forbidden \\
Inference temperature & $T=0$ \\
\bottomrule
\end{tabularx}
\caption{ReDNA settings.}
\label{tab:redna-hyperparams}
\end{table}


\subsection{Robustness to Decoding Stochasticity}
\label{app:robustness}

The main results in Table~\ref{tab:main_results} use greedy decoding. To verify that the qualitative claims of \S\ref{sec:main_results} are not artifacts of a single deterministic pass, we re-run every (model, method) cell on the four stronger-variant models at temperature $T=0.7$, three independent runs each. All other components of the protocol---scenarios, four-attempt evaluation, forbidden-finish-action diversity constraint, step budget, working-memory cap, and the LLM-as-a-judge---are kept identical to \S\ref{sec:setup}.

Table~\ref{tab:robustness} reports mean\,$\pm$\,sample standard deviation across the three runs. The two main-text directions hold under stochastic decoding: ReDNA achieves the highest Path Discovery on all four models, with gaps to the next-best method that exceed one standard deviation in every case; and ReDNA produces the highest Elaboration on all four models. Run-to-run variance is small in absolute terms---under 6 percentage points on Path Discovery and under 0.10 on action-level metrics in every cell---so the conclusions in \S\ref{sec:main_results} that rest on gaps larger than this scale are not driven by stochasticity.

\begin{table*}[t]
\centering
\footnotesize
\setlength{\tabcolsep}{6pt}
\renewcommand{\arraystretch}{1.1}
\begin{tabular}{ll ccccc}
\toprule
\textbf{Model} & \textbf{Method} & \textbf{Path \%} & \textbf{Avg Step} & \textbf{Originality} & \textbf{Elaboration} & \textbf{Groundedness} \\
\midrule
\multirow{4}{*}{GPT-5.4} & Base & 47.0\,$\pm$\,5.8 & 29.8\,$\pm$\,1.3 & 2.775\,$\pm$\,0.021 & 2.829\,$\pm$\,0.030 & 3.519\,$\pm$\,0.046 \\
 & Self-Refine & 45.8\,$\pm$\,2.7 & 31.0\,$\pm$\,4.6 & 2.765\,$\pm$\,0.054 & 2.737\,$\pm$\,0.011 & 3.490\,$\pm$\,0.064 \\
 & EscapeAgent & 61.3\,$\pm$\,2.7 & 38.7\,$\pm$\,1.6 & 2.441\,$\pm$\,0.024 & 2.919\,$\pm$\,0.067 & 3.573\,$\pm$\,0.012 \\
 & ReDNA & 64.9\,$\pm$\,2.7 & 40.8\,$\pm$\,2.6 & 2.713\,$\pm$\,0.048 & 3.304\,$\pm$\,0.030 & 3.715\,$\pm$\,0.069 \\
\midrule
\multirow{4}{*}{Claude Sonnet 4.6} & Base & 50.0\,$\pm$\,3.1 & 29.0\,$\pm$\,1.4 & 2.664\,$\pm$\,0.003 & 3.051\,$\pm$\,0.033 & 3.346\,$\pm$\,0.037 \\
 & Self-Refine & 48.8\,$\pm$\,5.8 & 21.3\,$\pm$\,1.1 & 2.739\,$\pm$\,0.020 & 3.072\,$\pm$\,0.019 & 3.377\,$\pm$\,0.098 \\
 & EscapeAgent & 55.9\,$\pm$\,3.7 & 37.9\,$\pm$\,3.0 & 2.475\,$\pm$\,0.031 & 2.813\,$\pm$\,0.006 & 3.323\,$\pm$\,0.048 \\
 & ReDNA & 61.3\,$\pm$\,5.4 & 28.9\,$\pm$\,1.0 & 2.618\,$\pm$\,0.017 & 3.092\,$\pm$\,0.029 & 3.461\,$\pm$\,0.045 \\
\midrule
\multirow{4}{*}{Llama-4-Maverick} & Base & 24.4\,$\pm$\,3.7 & 28.8\,$\pm$\,2.1 & 2.142\,$\pm$\,0.088 & 2.662\,$\pm$\,0.057 & 3.189\,$\pm$\,0.031 \\
 & Self-Refine & 23.8\,$\pm$\,1.0 & 33.4\,$\pm$\,0.9 & 2.301\,$\pm$\,0.061 & 2.713\,$\pm$\,0.015 & 3.137\,$\pm$\,0.061 \\
 & EscapeAgent & 29.8\,$\pm$\,2.7 & 43.3\,$\pm$\,3.3 & 2.213\,$\pm$\,0.058 & 2.628\,$\pm$\,0.057 & 3.159\,$\pm$\,0.018 \\
 & ReDNA & 53.0\,$\pm$\,2.7 & 45.1\,$\pm$\,2.5 & 2.330\,$\pm$\,0.008 & 3.074\,$\pm$\,0.061 & 3.088\,$\pm$\,0.042 \\
\midrule
\multirow{4}{*}{Qwen3-235B-A22B} & Base & 40.5\,$\pm$\,2.7 & 32.4\,$\pm$\,4.5 & 2.738\,$\pm$\,0.077 & 3.228\,$\pm$\,0.072 & 3.431\,$\pm$\,0.013 \\
 & Self-Refine & 41.7\,$\pm$\,2.1 & 33.3\,$\pm$\,0.2 & 2.696\,$\pm$\,0.068 & 3.203\,$\pm$\,0.055 & 3.459\,$\pm$\,0.032 \\
 & EscapeAgent & 48.2\,$\pm$\,3.1 & 40.2\,$\pm$\,2.7 & 2.695\,$\pm$\,0.044 & 3.197\,$\pm$\,0.050 & 3.459\,$\pm$\,0.039 \\
 & ReDNA & 56.0\,$\pm$\,1.0 & 35.8\,$\pm$\,2.6 & 2.692\,$\pm$\,0.055 & 3.424\,$\pm$\,0.055 & 3.375\,$\pm$\,0.045 \\
\bottomrule
\end{tabular}
\caption{Robustness study at $T{=}0.7$, three runs per cell; values are mean\,$\pm$\,sample standard deviation.}
\label{tab:robustness}
\end{table*}

\subsection{AUT Implementation Details} 
\label{app:aut-details}
We evaluate both Base and ReDNA on the AUT using GPT-5.4 with sampling temperature $T=0.7$, averaged over three independent runs per condition.
Base produces all uses in a single call. ReDNA follows the input-separation structure of \S\ref{sec:method}: the Diverge call generates candidate uses conditioned only on the object, and the Narrowing call selects and elaborates them under the creativity criterion, using previously generated uses from earlier runs as
the non-interactive analog of failure memory. Originality and Elaboration are scored on a 1--5 scale following the rubric of~\citep{guilford1967creativity}.

\paragraph{Use of Large Language Models.}
We write the manuscript ourselves. We use an LLM (ChatGPT) solely for language refinement, including style, clarity, and grammar, and do not use it for research ideation or analysis.

\section{MUTATE Benchmark Details}
\label{app:mutate-details}

\subsection{YAML Scenario Structure}

We store each benchmark instance as a YAML file. A file represents one interactive scenario and is encoded as a list of scenes. The first scene also contains the scenario-level \texttt{objective}, which we show to the agent as its goal. Listing~\ref{lst:yaml-scenario} shows the schema skeleton.

\begin{lstlisting}[
  style=appendixprompt,
  basicstyle=\footnotesize\ttfamily,
  caption={YAML scenario skeleton.},
  label={lst:yaml-scenario}
]
- name: scene_name
  objective: "Scenario goal."   # first scene only
  desc: "Textual scene description."
  visible: true                 # optional; default true
  scene_relations:
    Move label: other_scene_name

  items:
  - position: "Item location."
    item:
      name: item_name
      visible: true             # optional; default true
      interactable: true        # optional; default true
      states:
      - desc: "Item state description."
        neg_reward: "Failure feedback."
        transitions:
        - wait_for: [apply, tool_name]
          trigger: [change_state, 1]
          reward: "Success feedback."

  tools:
  - position: "Tool location."
    tool:
      name: tool_name
      visible: true
      states:
      - desc: "Tool description."
        apply_to: [target_item_name]
        wait_for: [ingredient_tool_name]
\end{lstlisting}

We treat a scenario as a stateful text environment. The agent occupies one scene at a time and observes the current scene description, visible items, visible tools, nearby scenes, and tools already collected in its bag. The \texttt{scene\_relations} field defines the navigation graph: each key is the movement label exposed to the agent, and each value is the destination scene.

We divide objects into items and tools. Items are scene objects that can be inspected or manipulated but are not normally collected. Tools are collectable objects; when the agent executes \texttt{click(tool)}, the tool moves into the bag and can later be used in \texttt{apply(tool, item)} or \texttt{craft(tool\_a, tool\_b)} actions.

Each item or tool may have multiple states. The current state determines its visible description and available interactions. Item states may define transitions, where \texttt{wait\_for} specifies the action pattern, \texttt{trigger} specifies the world-state update, and \texttt{reward} gives
the environment feedback. If no transition matches, the environment returns \texttt{neg\_reward} when provided, or a generic failure message otherwise.

The action space is fixed: \texttt{click(x)} inspects an item or collects a visible tool; \texttt{apply(t, x)} applies a collected tool to an item; \texttt{craft(t1, t2)} combines two collected tools; \texttt{input(s, x)} enters a string into an item; and \texttt{move(scene)} moves to an adjacent scene.

The \texttt{trigger} field supports state changes such as changing the current item state, changing another item or tool state, toggling visibility, toggling interactability, or converting an item into a collectable tool. Successful terminal solutions are marked textually in the transition reward with \texttt{GAME END!}. Some two-phase scenarios contain intermediate rewards marked with \texttt{CHECKPOINT!}; we use them to separate Phase 1 and Phase 2 solution discovery. Separate evaluation metadata files specify valid solution path IDs, trigger actions, phases, and principal types.

\subsection{Solution Path Catalog}

Tables~\ref{tab:path-catalog-twophase} and~\ref{tab:path-catalog-single} list the gold path catalog. Each phase contains one A path, one B path, and two C paths unless the scenario intentionally leaves a path unavailable.

\begin{table*}[t]
\centering
\small
\setlength{\tabcolsep}{3.5pt}
\renewcommand{\arraystretch}{1.1}
\begin{tabularx}{\textwidth}{p{0.13\textwidth} p{0.12\textwidth} X X X X}
\toprule
\textbf{Scenario} & \textbf{Phase} & \textbf{A} & \textbf{B} & \textbf{C1} & \textbf{C2} \\
\midrule
House Heist & Safe Access & Break the safe with the pipe head. & Recover a wine-glass fingerprint with adhesive tape. & Cut the power cable. & Decode binary override \texttt{10010110}. \\
 & Room Escape & Use the door key. & Break the window pane with scissors. & Use the curtain hook and chimney opening. & Decode reversed door code \texttt{2518}. \\
\midrule
Desert Van Restart & Tire Mobility & Replace the wheel. & Patch and reinflate the tire. & No tire-side C1 path. & Call desert recovery service. \\
 & Engine Restart & Replace the hose. & Seal the hose with gum or tape. & Use a donor engine part. & Call desert recovery service. \\
\midrule
Laundry Machine 7 & Basement Access & Move machine 2 with the dolly. & Open the vent grate with a coin. & Loosen the rusted handle with detergent. & Infer passcode \texttt{0411}. \\
 & Machine Stop & Input timer code \texttt{000}. & Damage the drain hose with the fire axe. & Map digit-7 strokes to \texttt{abceg}. & Select service option \texttt{3}. \\
\midrule
Sinking Ship & Deck Access & Infer door code \texttt{1704}. & Open the maintenance hatch. & Use the duct route. & Restore the service lift. \\
 & Rescue/Escape & Fire the flare gun. & Repair and inflate the lifeboat. & Restore radio channel \texttt{16}. & Operate winch/davit release. \\
\bottomrule
\end{tabularx}
\caption{Two-phase solution paths.}
\label{tab:path-catalog-twophase}
\end{table*}

\begin{table*}[t]
\centering
\small
\setlength{\tabcolsep}{3.5pt}
\renewcommand{\arraystretch}{1.1}
\begin{tabularx}{\textwidth}{p{0.15\textwidth} X X X X}
\toprule
\textbf{Scenario} & \textbf{A} & \textbf{B} & \textbf{C1} & \textbf{C2} \\
\midrule
Beehive Removal & Wear protection and remove the nest. & Calm bees with newspaper smoke. & Restore the flyer number and call experts. & Use floral lure toward the garden tree. \\
\midrule
Missed Call & Cover the speaker with a comforter. & Stop the laptop with the bookend. & Infer the laptop password. & Close the closet door. \\
\midrule
Pipe Leak & Tighten the pipe joint. & Seal the pipe with a rubber patch. & Input pressure code \texttt{1225}. & Open the manhole to drain water. \\
\midrule
Space Station Fire & Open bulkhead with code \texttt{1969}. & Use extinguisher propulsion. & Use the pressure dump lever. & Use the zero-g crawlway. \\
\midrule
Underground Mine & Use sedative jerky. & Repel the dog with sulfur odor. & Input voice keyword \texttt{blaze}. & Burst the ceiling pipe with explosives. \\
\midrule
Zombie Lab & Use the cola noise lure. & Build an infected scent disguise. & Open the vent route. & Craft a firebomb. \\
\bottomrule
\end{tabularx}
\caption{Single-phase solution paths.}
\label{tab:path-catalog-single}
\end{table*}

\subsection{Path Discovery per Scenario}

We report per-path discovery with the model-method entries for which per-path coverage is available. Table~\ref{tab:path-discovery-twophase} reports two-phase scenarios, and Table~\ref{tab:path-discovery-single} reports single-phase scenarios. \okmark{} indicates that the path is discovered, and \xmarktab{} indicates that it is not discovered.

\begin{table*}[t]
\centering
\small
\setlength{\tabcolsep}{3pt}
\renewcommand{\arraystretch}{1.02}
\begin{tabular*}{\textwidth}{@{\extracolsep{\fill}} l l cccccccc}
\toprule
\textbf{Scenario} & \textbf{Method} & \textbf{P1\_A} & \textbf{P1\_B} & \textbf{P1\_C1} & \textbf{P1\_C2} & \textbf{P2\_A} & \textbf{P2\_B} & \textbf{P2\_C1} & \textbf{P2\_C2} \\
\midrule
House Heist & Claude Base & X & O & O & X & X & O & X & O \\
House Heist & Claude ReDNA & X & O & O & X & O & O & X & X \\
House Heist & GPT Base & X & O & O & X & X & O & O & X \\
House Heist & GPT ReDNA & O & X & O & X & X & O & O & X \\
House Heist & Llama Base & X & O & X & X & O & X & X & X \\
House Heist & Llama ReDNA & X & X & O & X & X & X & X & X \\
House Heist & Qwen Base & X & X & X & X & X & X & X & X \\
House Heist & Qwen ReDNA & X & O & O & X & O & X & O & X \\
\midrule
Desert Van & Claude Base & O & O & X & X & O & O & O & X \\
Desert Van & Claude ReDNA & O & O & X & X & O & O & O & X \\
Desert Van & GPT Base & O & O & X & X & O & X & O & X \\
Desert Van & GPT ReDNA & O & O & X & X & O & O & O & X \\
Desert Van & Llama Base & X & X & X & X & X & X & X & X \\
Desert Van & Llama ReDNA & O & X & X & X & O & O & X & X \\
Desert Van & Qwen Base & X & O & X & X & X & X & X & X \\
Desert Van & Qwen ReDNA & O & O & X & X & O & O & X & X \\
\midrule
Laundry 7 & Claude Base & X & O & O & X & X & X & O & X \\
Laundry 7 & Claude ReDNA & X & O & O & O & O & X & O & X \\
Laundry 7 & GPT Base & O & O & X & O & X & X & X & X \\
Laundry 7 & GPT ReDNA & X & O & O & X & X & O & X & O \\
Laundry 7 & Llama Base & X & O & X & X & O & X & X & X \\
Laundry 7 & Llama ReDNA & O & O & O & X & O & X & X & X \\
Laundry 7 & Qwen Base & X & O & X & O & X & X & X & X \\
Laundry 7 & Qwen ReDNA & O & O & O & O & O & X & X & O \\
\midrule
Sinking Ship & Claude Base & X & O & O & O & X & O & X & X \\
Sinking Ship & Claude ReDNA & X & O & O & O & X & O & O & O \\
Sinking Ship & GPT Base & X & O & O & O & X & O & X & O \\
Sinking Ship & GPT ReDNA & X & O & O & O & X & O & O & O \\
Sinking Ship & Llama Base & X & O & X & X & X & X & X & X \\
Sinking Ship & Llama ReDNA & X & O & O & O & X & O & O & O \\
Sinking Ship & Qwen Base & X & O & O & O & X & O & O & X \\
Sinking Ship & Qwen ReDNA & O & O & O & O & X & O & O & O \\
\bottomrule
\end{tabular*}
\caption{Two-phase path discovery.}
\label{tab:path-discovery-twophase}
\end{table*}

\begin{table*}[t]
\centering
\small
\setlength{\tabcolsep}{4pt}
\renewcommand{\arraystretch}{1.02}
\begin{tabular*}{\textwidth}{@{\extracolsep{\fill}} l l cccc}
\toprule
\textbf{Scenario} & \textbf{Method} & \textbf{A} & \textbf{B} & \textbf{C1} & \textbf{C2} \\
\midrule
Beehive & Claude Base & O & O & X & O \\
Beehive & Claude ReDNA & O & O & X & O \\
Beehive & GPT Base & O & X & X & X \\
Beehive & GPT ReDNA & O & O & O & O \\
Beehive & Llama Base & O & X & X & X \\
Beehive & Llama ReDNA & O & O & X & X \\
Beehive & Qwen Base & O & X & O & X \\
Beehive & Qwen ReDNA & O & X & O & X \\
\midrule
Missed Call & Claude Base & X & O & X & O \\
Missed Call & Claude ReDNA & O & O & X & O \\
Missed Call & GPT Base & X & X & X & O \\
Missed Call & GPT ReDNA & X & X & X & O \\
Missed Call & Llama Base & X & X & X & X \\
Missed Call & Llama ReDNA & O & X & X & O \\
Missed Call & Qwen Base & X & X & X & O \\
Missed Call & Qwen ReDNA & O & X & X & O \\
\midrule
Pipe Leak & Claude Base & O & X & X & X \\
Pipe Leak & Claude ReDNA & O & O & O & O \\
Pipe Leak & GPT Base & O & O & O & X \\
Pipe Leak & GPT ReDNA & O & O & O & O \\
Pipe Leak & Llama Base & O & O & X & O \\
Pipe Leak & Llama ReDNA & O & O & X & O \\
Pipe Leak & Qwen Base & O & O & O & O \\
Pipe Leak & Qwen ReDNA & O & O & O & X \\
\midrule
Space Fire & Claude Base & O & X & O & O \\
Space Fire & Claude ReDNA & O & O & O & O \\
Space Fire & GPT Base & X & O & O & O \\
Space Fire & GPT ReDNA & O & O & O & X \\
Space Fire & Llama Base & X & O & O & O \\
Space Fire & Llama ReDNA & O & O & O & O \\
Space Fire & Qwen Base & O & X & O & O \\
Space Fire & Qwen ReDNA & O & X & O & O \\
\midrule
Mine & Claude Base & O & X & O & X \\
Mine & Claude ReDNA & O & X & O & X \\
Mine & GPT Base & O & X & O & X \\
Mine & GPT ReDNA & O & X & O & O \\
Mine & Llama Base & O & X & X & X \\
Mine & Llama ReDNA & O & X & X & X \\
Mine & Qwen Base & O & X & X & X \\
Mine & Qwen ReDNA & O & X & O & X \\
\midrule
Zombie Lab & Claude Base & O & X & O & X \\
Zombie Lab & Claude ReDNA & O & O & O & O \\
Zombie Lab & GPT Base & O & X & X & X \\
Zombie Lab & GPT ReDNA & O & X & O & O \\
Zombie Lab & Llama Base & X & X & X & X \\
Zombie Lab & Llama ReDNA & O & X & O & X \\
Zombie Lab & Qwen Base & O & X & O & O \\
Zombie Lab & Qwen ReDNA & X & O & O & O \\
\bottomrule
\end{tabular*}
\caption{Single-phase path discovery.}
\label{tab:path-discovery-single}
\end{table*}

\subsection{Benchmark UI for Human Evaluation}

\begin{figure*}[t]
\centering
\begin{subfigure}{0.48\textwidth}
\centering
\includegraphics[width=\linewidth]{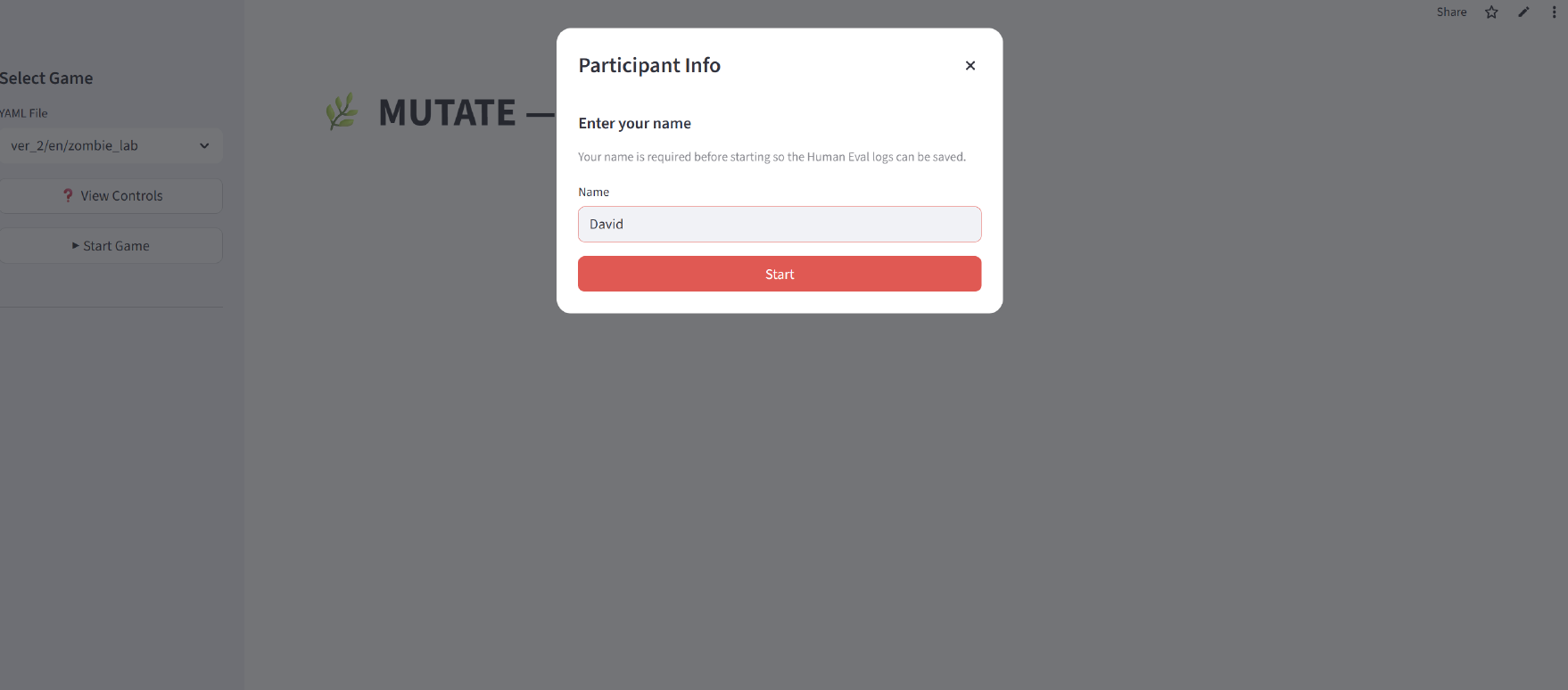}
\caption{Login.}
\end{subfigure}
\hfill
\begin{subfigure}{0.48\textwidth}
\centering
\includegraphics[width=\linewidth]{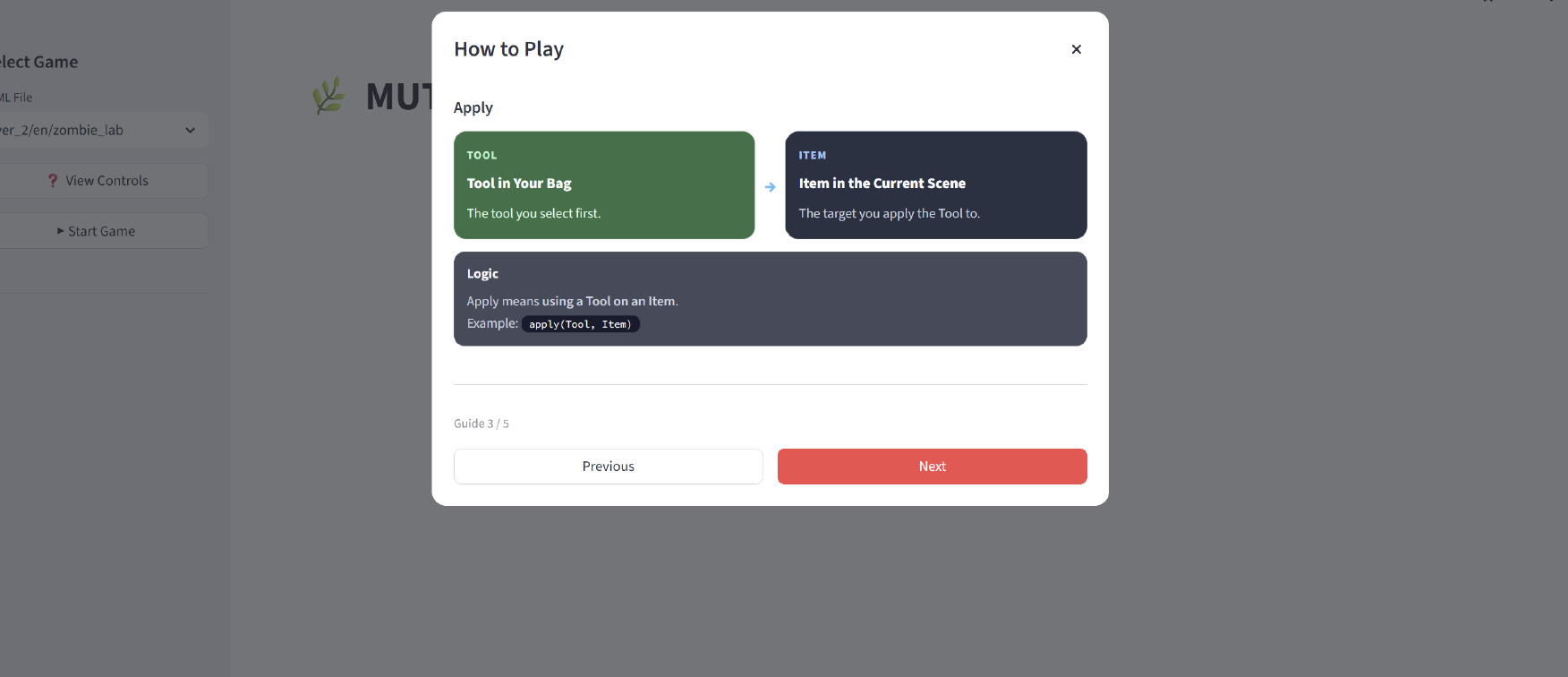}
\caption{Tutorial.}
\end{subfigure}
\vspace{4pt}

\begin{subfigure}{0.48\textwidth}
\centering
\includegraphics[width=\linewidth]{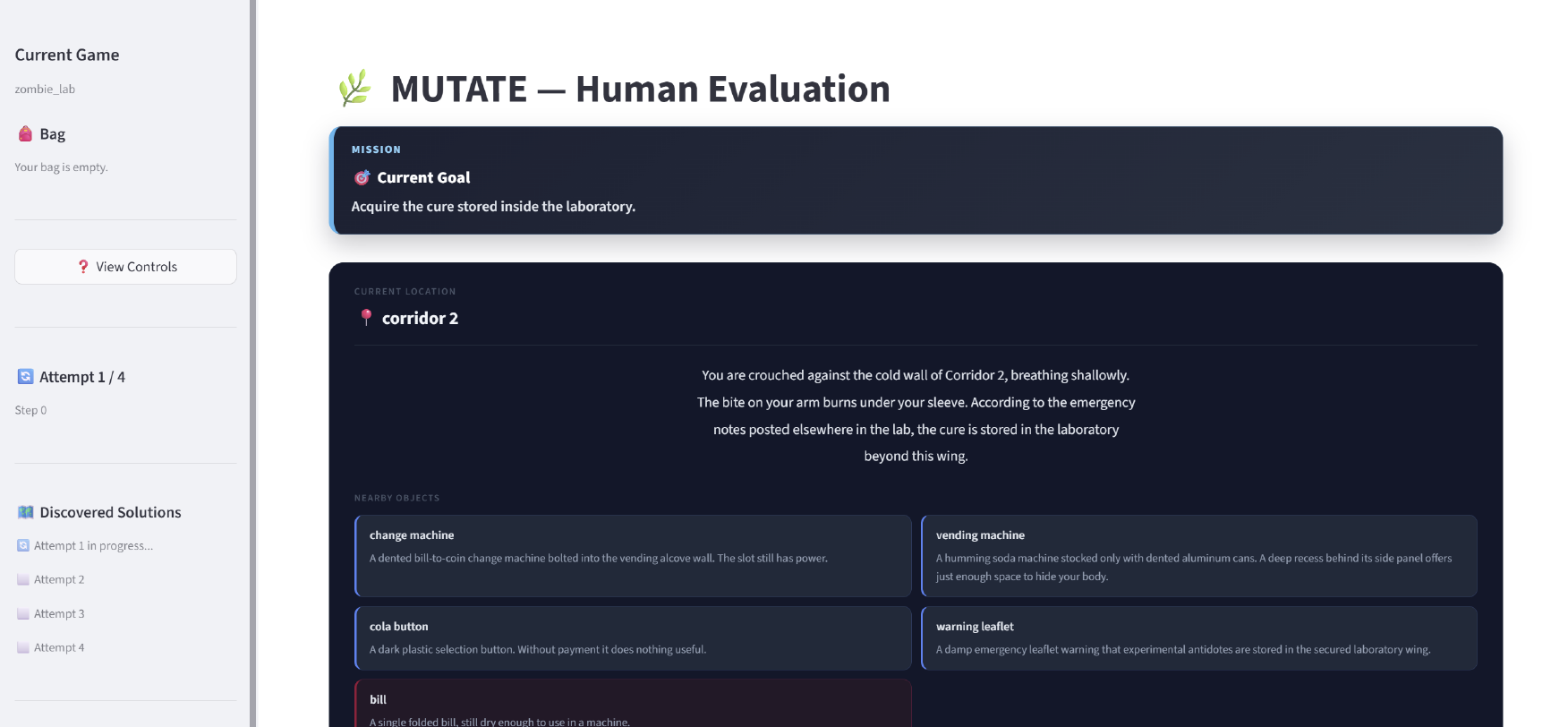}
\caption{Main view.}
\end{subfigure}
\hfill
\begin{subfigure}{0.48\textwidth}
\centering
\includegraphics[width=\linewidth]{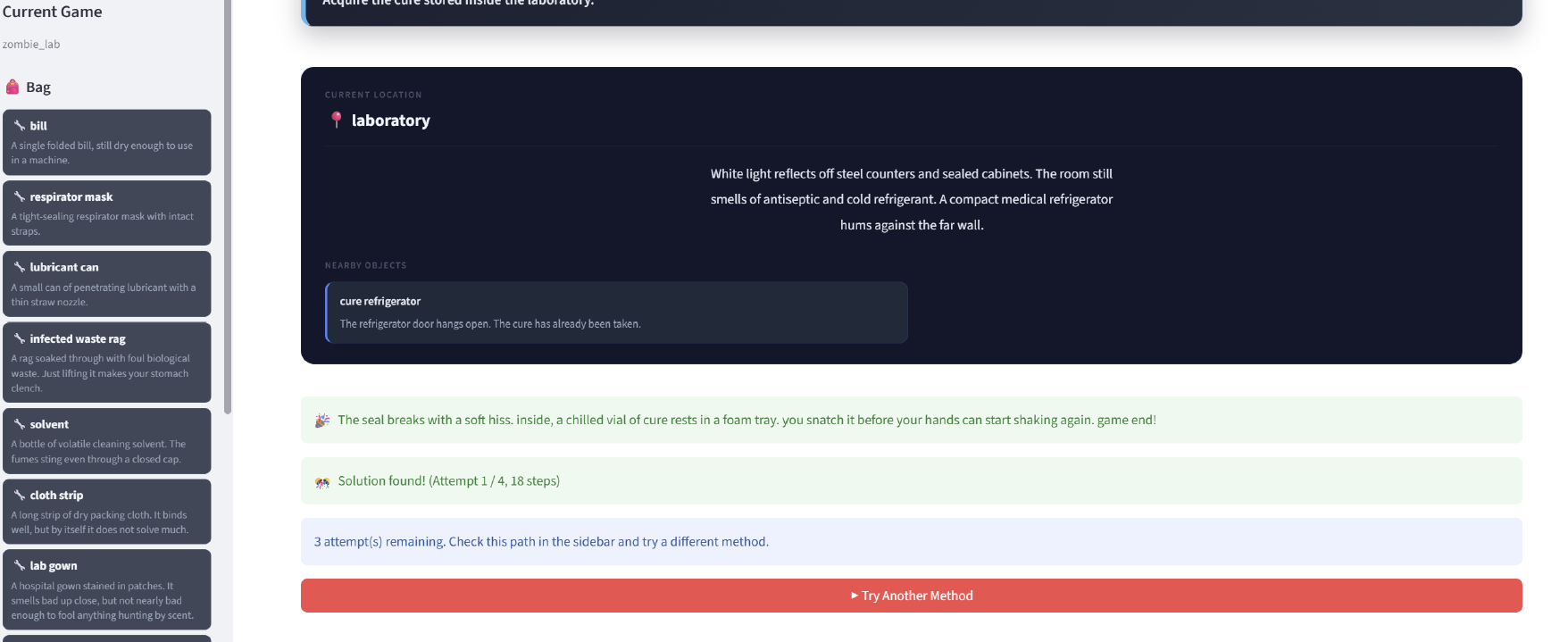}
\caption{Scoring view.}
\end{subfigure}
\vspace{4pt}

\begin{subfigure}{0.48\textwidth}
\centering
\includegraphics[width=\linewidth]{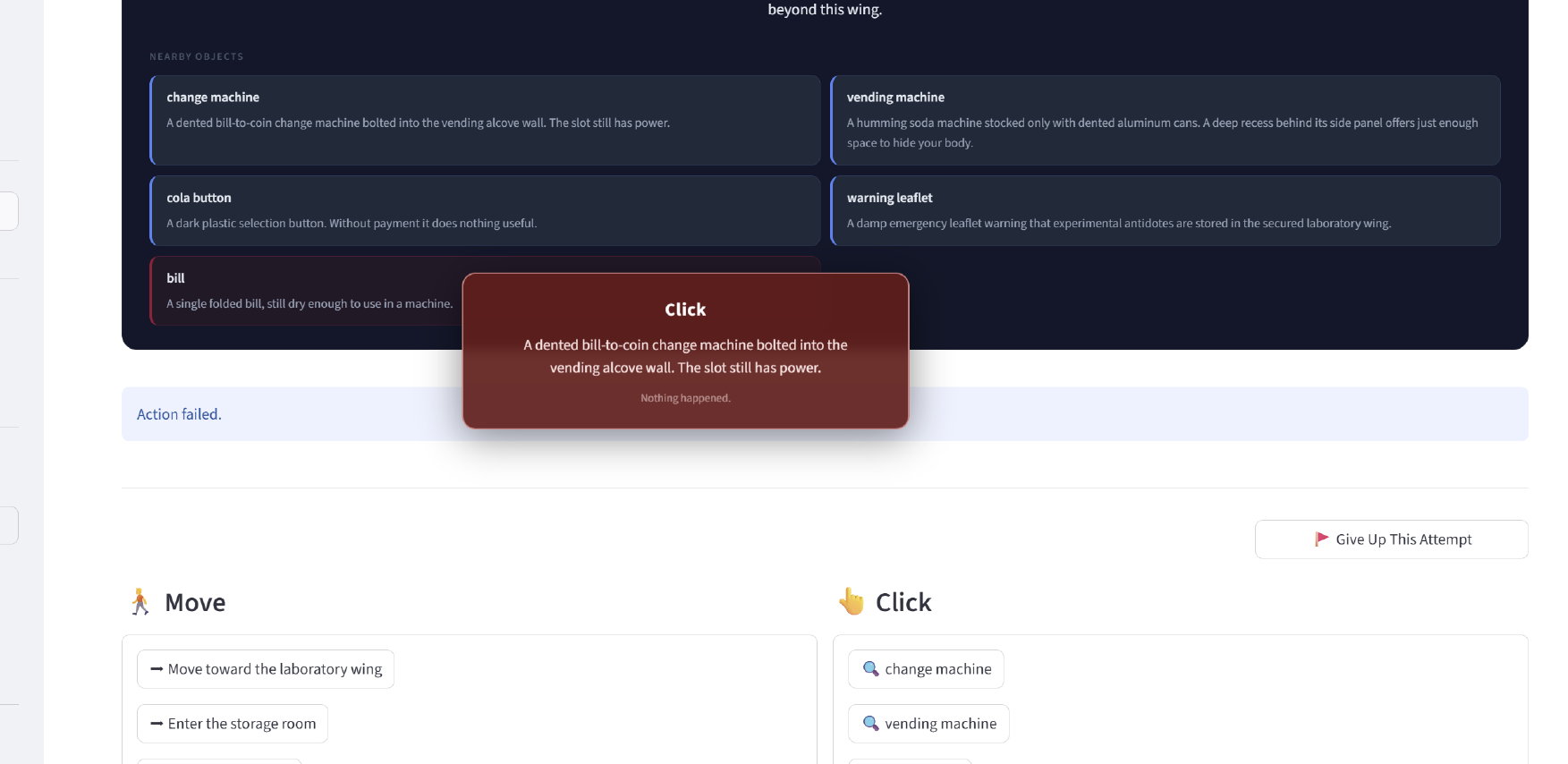}
\caption{Context view.}
\end{subfigure}
\hfill
\begin{subfigure}{0.48\textwidth}
\centering
\includegraphics[width=\linewidth]{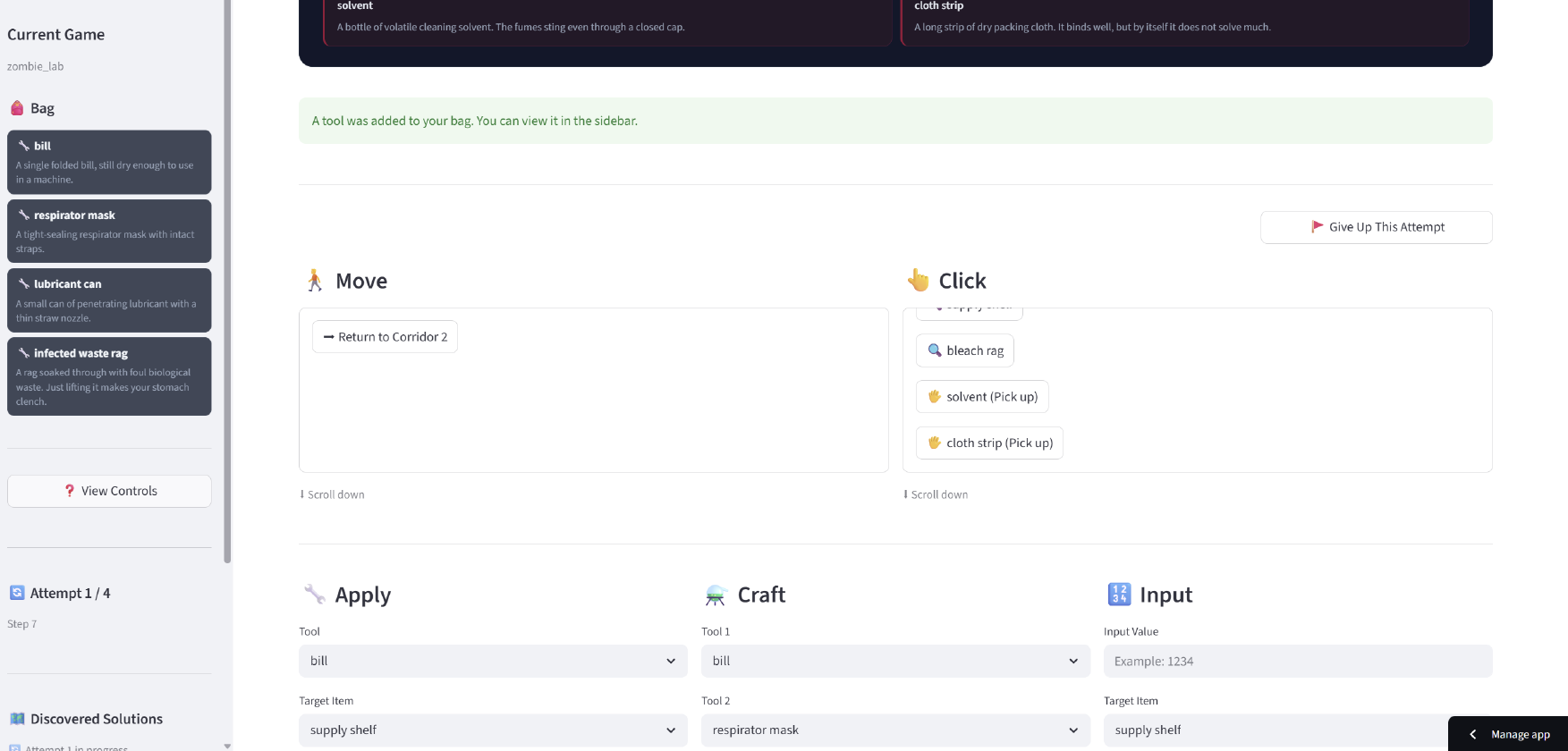}
\caption{Bag view.}
\end{subfigure}
\caption{Human evaluation UI.}
\label{fig:eval-ui}
\end{figure*}

We implement the human evaluation interface as a web UI. Annotators log in, read the tutorial, then score sampled Thought-Action records against the displayed environment state, action history, current bag contents, and scenario objective. Figure~\ref{fig:eval-ui} shows the login, tutorial, scoring, context, and bag views.

\clearpage
\onecolumn
\raggedbottom
\ifcsname nolinenumbers\endcsname
  \nolinenumbers
\fi

\tcbset{promptBoxBase/.style={%
  enhanced,
  width=0.92\linewidth,
  colback=black!12, colframe=black!60,
  boxrule=0.8pt, arc=2pt,
  left=1em, right=1em, top=0.7em, bottom=0.7em,
  listing only,
  listing options={%
    basicstyle=\scriptsize\ttfamily,
    breaklines=true,
    breakatwhitespace=false,
    columns=fullflexible,
    keepspaces=true,
    showstringspaces=false
  },
  before skip=0.8em,
  after skip=0.35em,
}}

\newtcblisting{PromptBox}[1][]{%
  promptBoxBase,
  breakable,
  #1}

\newcommand{\PromptCaption}[2]{%
  \refstepcounter{figure}%
  \if\relax\detokenize{#1}\relax\else\label{#1}\fi%
  \begin{center}
    \parbox{0.92\linewidth}{\centering\small\textbf{Figure~\thefigure:} #2\par}
  \end{center}
  \vspace{0.8\baselineskip}%
}

\section{Prompts}
\label{appn:prompt}

This section lists the full prompt templates used for the Base agent,
EscapeAgent, Self-Refine, and ReDNA.

\begin{center}
\begin{PromptBox}
You are interacting with a simulated environment. Your objective is to accomplish the scenario's functional goal (e.g., stabilize a system, repair something, restore operation).
IMPORTANT: There may be MULTIPLE valid ways to achieve the goal. Try different approaches and discover alternative solution paths when possible.

There are three types of interactives:
- items: interactable objects in the CURRENT SCENE
- tools: usable tools already collected IN YOUR BAG
- scenes: nearby scenes you can move to

You can perform ONE of the following actions:
- click(<interactable item or visible tool>): Examine an item in the current scene, or collect a visible tool from the current scene into your bag.
- apply(<tool in your bag>, <interactable item in the current scene>): Apply a tool in your bag to an ITEM in the current scene.
- input(string, <interactable item in the current scene>): Input a string to an item in the current scene.
- move(<interactable scene>): Move to a nearby scene.
- craft(<base tool in your bag>, <ingredient tool in your bag>): Apply the ingredient tool onto the base tool to upgrade or modify the base tool. The FIRST argument is the base tool. The SECOND argument is the ingredient applied onto it. Order matters: craft(a, b) is NOT craft(b, a).

Examples:
click(microwave)
click(wrench)
apply(key, silver chest)
craft(controller, battery)
input(c79a1, combination lock)
move(Go to operation room)

CRITICAL OPERATIONAL RULES:
1) You may ONLY use names explicitly listed in Possible Actions.
2) A visible tool in the scene is NOT in your bag yet. If you want to use it later, click it first.
3) craft can ONLY be used between TWO TOOLS already in your bag. The FIRST argument is the base tool. The SECOND argument is the ingredient tool.
4) apply can ONLY target an INTERACTABLE ITEM IN THE CURRENT SCENE. The second argument of apply() must appear under Possible Actions items. If the target is a tool, use craft().
5) If you are considering craft or apply, first check whether both required tools are already in your bag, and whether the target is an item rather than a tool.
6) If an action fails, do not repeat the same failed hypothesis immediately. Change the target, collect missing tools, move, or try a different mechanism.
7) Prefer actions that increase future options: explore scenes, click unexplored items, collect relevant tools, then apply or craft from observed evidence.
8) Because there can be multiple solutions, do not assume there is only one correct next step.
9) After craft(A, B) SUCCEEDS, do NOT attempt craft(A, B) or craft(B, A) again. The ingredient B is consumed on success.

ANTI-LOOP RULES:
- If an action fails ONCE, do NOT repeat the exact same action.
- NEVER use apply(X, Y) where Y is a tool in your bag or a visible tool in the scene. Use craft() for tool-on-tool operations.
- After craft(A, B) succeeds, never repeat craft(A, B).
- If you repeat any action without progress, move to a different scene or approach.

Before choosing an action, silently verify:
- Is the target in the current scene's item list?
- Is the thing I want to use already in my bag?
- For apply(X, Y): Is Y listed under items in the current scene?
- For craft(A, B): Have I already done this craft successfully?
- Have I tried this exact action before and failed?

Please respond in two lines.
Thought: ...
Action: ...
\end{PromptBox}
\end{center}
\PromptCaption{fig:prompt-shared-system}{Prompt 1. Shared System Prompt (Base / EscapeAgent / Self-Refine).}
\label{fig:prompt-shared-system-actions}
\label{fig:prompt-shared-system-rules}

\clearpage
\begin{center}
\begin{PromptBox}
<RECENT_HISTORY>

YOUR OBJECTIVE: <GAME_OBJECTIVE>

Now you need to act on [Step <STEP>]
Your current position is: <POSITION>.

<SCENE_DESCRIPTION>
<POSSIBLE_ACTIONS>
<TOOLS_IN_BAG>

NOTE:
- When you use move, you MUST use exactly one listed scene name.
- Do NOT include extra description such as ": It leads to ...".
- Valid scene names: <VALID_SCENE_NAMES>
- Example: move(<VALID_SCENE_NAME>)

IMPORTANT EXPLORATION RULES:
- If an action fails and nothing changes, do NOT repeat it.
- If the environment gave a MEANINGFUL response, stay with that target and try a DIFFERENT tool or approach.
- If the environment gave NO response ("Nothing happens"), abandon that specific action.
- Avoid repeating the same tool-target interaction.
- Explore different strategies: different tools, crafting, different objects, or another scene.

Your goal is creative exploration, not repeating identical attempts.

<DIVERSITY_CONSTRAINT_IF_ANY>

Your Response:
\end{PromptBox}
\end{center}
\PromptCaption{fig:prompt-base-user}{Prompt 2. Base User Prompt Template (Base).}

\begin{center}
\begin{PromptBox}
DIVERSITY CONSTRAINT (VERY IMPORTANT):
- You are running multiple trials. You MUST discover a DIFFERENT solution each trial.
- You MUST NOT finish the game using any of the following forbidden finish actions:
  * <FORBIDDEN_FINISH_ACTION>
- If you think one of them would solve it, you must intentionally pursue an alternative approach.
\end{PromptBox}
\end{center}
\PromptCaption{fig:prompt-diversity}{Prompt 3. Diversity Constraint (All methods).}

\begin{center}
\begin{PromptBox}
You are currently exploring the scene freely. You should try explore new scenes, interact with the items through click, input or apply actions, and try crafting new tools:
- If there's still <interactable items> you haven't tried any action to interact with, you should try 'click' them first.
- Otherwise, explore other new <interactable scene> you haven't been to, or going back to parent scene.
- Follow the rules in Possible Actions and system prompt to give a valid action and thought.
Do not repeat actions in history and previous steps. Your Response:
\end{PromptBox}
\end{center}
\PromptCaption{fig:prompt-free-exploration}{Prompt 4. Free Exploration (EscapeAgent).}
\clearpage
\begin{center}
\begin{PromptBox}
You are a helpful agent to reflect on your action and environment response, and then maintain a task list with solving suggestions.
The role of this task list is that there are some tasks you currently cannot solve with the tools at hand, but you think you may need to solve them later, so write them down with some suggestions and hints for your future reference.

After analyzing your current action and the response from the environment, you should give an action to maintain the task list: <function_list>
<param_explain>
For instance, valid task list maintaining action may be: <use_example>.

Reflection functions:
update(updated_feedback)
The parameter should retain the original feedback and add one new hindsight from the current action.

new(task_name, feedback)
The first parameter is a brief task name. The second parameter is what you have to do to solve this task.

delete(index)
If you choose delete, the first parameter is the index of the completed or useless task.

none()
If you choose none, do not give any parameter.
\end{PromptBox}
\end{center}
\PromptCaption{fig:prompt-reflection-system}{Prompt 5. Reflection System (EscapeAgent).}

\begin{center}
\begin{PromptBox}
Your current position:
<POSITION>

<SCENE_DESCRIPTION>

<POSSIBLE_ACTIONS>

Your action: <ACTION>

Response from the environment:
<ENVIRONMENT_RESPONSE>

Now please make an action call to maintain the task list in one line. Follow the system instruction to extract hint and fill in the parameter for the function call.

Your Response:
\end{PromptBox}
\end{center}
\PromptCaption{fig:prompt-reflection-user}{Prompt 6. Reflection User (EscapeAgent).}
\clearpage

\begin{center}
\begin{PromptBox}
You have to use your creativity to figure out the use of the tool you have just collected.

There are generally two ways about how to use the tool:
1. Combine this tool with another one in your bag to craft a new tool: craft(<collected tool>, <applicable tool>).
2. Apply this tool to a target item in a task: apply(<collected tool>, Target Item in a task).

Hints:
1. Pay attention to the task and tool descriptions. Find the connection between them.
2. In Thought, explicitly consider bag items for crafting and task-list targets for applying.
3. In Actions, give zero to multiple craft/apply calls. For apply, give the task index and justify why the tool may solve the task.

User template:
You have just collected a new tool:
<collected tool>: <TOOL_NAME>
Description: <TOOL_DESCRIPTION>

Other tools in your bag:
<TOOLS_IN_BAG>

Tasks waiting to be solved:
<TASK_LIST>

Please follow the system prompt to output your Thought and Actions. Analyze thoroughly and be bold to propose plausible craft and apply actions.
\end{PromptBox}
\end{center}
\PromptCaption{fig:prompt-forethought-tool}{Prompt 7. Forethought for New Tool (EscapeAgent).}

\begin{center}
\begin{PromptBox}
You have to use your creativity to figure out if you could use any tools you have now to solve a new task you have just discovered.

There are generally three ways to solve a task:
1. Click the target item.
2. Apply a bag tool to the target item.
3. Input a string to the target item.

Hints:
1. Pay attention to what the task needs. Always first try simple click if not done.
2. Examine tool descriptions and memory-pad hints to connect them to the task.
3. In Thought, consider click, apply, and input possibilities.
4. In Actions, give zero to multiple click/apply/input calls and justify each.

User template:
The current task:
[Task] Name: <TASK_NAME>, Target Item: <TARGET_ITEM>
<TASK_DESCRIPTION>

Tools in your bag:
<TOOLS_IN_BAG>

Hints from the memory pad:
<MEMORY_PAD>

Please follow the system prompt to output your Thought and Actions. Analyze thoroughly and be bold to propose plausible click, apply, and input actions.
\end{PromptBox}
\end{center}
\PromptCaption{fig:prompt-forethought-task}{Prompt 8. Forethought for New Task (EscapeAgent).}

\clearpage
\begin{center}
\begin{PromptBox}
SELF-REFLECT FEASIBILITY CHECK

Before choosing the final action, silently perform a step-wise feasibility check:
- Is the action physically feasible?
- Does it use only objects explicitly listed in Possible Actions or Tools in Bag?
- Is the action grammar valid?
- For apply(tool, item), is the tool already in the bag and is the target an item in the current scene?
- For craft(tool A, tool B), are both tools already in the bag?
- For move(scene), is the scene name one of the valid listed scene names?
- Does the action plausibly move toward the objective?
- Does it avoid repeating a clearly failed attempt from the history?

If your first candidate is invalid or weakly grounded, revise it before writing the final answer.
Do not show a separate verification section. Only output the final checked action.

Thought: ...
Action: ...
\end{PromptBox}
\end{center}
\PromptCaption{fig:prompt-self-refine}{Prompt 9. Self-Refine Suffix (Self-Refine).}

\begin{center}
\begin{PromptBox}
You are a one-shot feedback-guided creative reasoning module for a text environment.

The normal action policy has been temporarily replaced because the agent has accumulated feedback from failed interactions with the same target item.
Your job is to use a Divergent-Convergent process to choose the NEXT action.

Game action grammar:
- apply(<tool in your bag>, <interactable item in the current scene>)
- input(string, <interactable item in the current scene>)
- craft(<base tool in your bag>, <ingredient tool in your bag>)

Operational rules:
- Use only names explicitly listed in Possible Actions or Tools in Bag.
- apply() can only use a bag tool as the first argument and a current-scene item as the second argument.
- craft() can only combine two bag tools. Order matters.
- input() can only target a current-scene item.
- The final action must be apply(), input(), or craft(). Do not choose click() or move().

Think in two phases:
1. Divergent phase: generate diverse candidate mechanisms and action sketches. Do not choose the final action yet.
2. Convergent phase: apply the objective, valid actions, and accumulated failure evidence to choose exactly one valid next action.

Use only creative interaction mechanisms:
1. Try a new tool-based mechanism on the same target with apply().
2. Try an evidence-based input mechanism on the same target with input().
3. Craft tools to create a new capability that could address the target or objective.

Thought: ...
Action: ...
\end{PromptBox}
\end{center}
\PromptCaption{fig:prompt-redna-system}{Prompt 10. ReDNA System Prompt (ReDNA).}
\clearpage

\begin{center}
\begin{PromptBox}
FEEDBACK-GUIDED R/DN MODULE

Current step: <STEP>
Current position: <POSITION>
Feedback target item: <TARGET_ITEM>
Failure count on this target: <FAILURE_COUNT>
Non-click failure count on this target: <NON_CLICK_FAILURE_COUNT>

DIVERGENT PHASE INPUTS
Target item:
<TARGET_ITEM>

Current scene context:
<SCENE_DESCRIPTION>

Tools in Bag:
<TOOLS_IN_BAG>

DIVERGENT PHASE
Generate 2-4 substantially different candidate mechanisms using apply(), input(), or craft().
Consider:
- apply(tool, target item): apply a bag tool to the target item in a non-obvious way.
- input(string, target item): try an evidence-based code, word, label, or sequence.
- craft(tool A, tool B): combine two bag tools before returning to the target.
Do not propose click() or move().

CONVERGENT PHASE INPUTS
Objective:
<GAME_OBJECTIVE>

Accumulated failed attempts:
<FAILED_ATTEMPTS_WITH_RESPONSES>

Possible Actions:
<POSSIBLE_ACTIONS>

<DIVERSITY_CONSTRAINT_IF_ANY>

CONVERGENT PHASE
Choose exactly one valid next action.
Rules:
- Do not repeat any failed action exactly.
- Reject candidates contradicted by failure responses.
- Use accumulated feedback to change the failed mechanism.
- Choose a valid, non-repeated apply(), input(), or craft() action.
- If choosing input(), the string should be evidence-based.
- If choosing craft(), explain why the two tools combine into a useful capability.
- The final Action must start with apply(, input(, or craft(.

Before writing the final response, explain what the latest failure implies, compare 2-4 concrete candidates, and choose the final action because it addresses what the failed action lacked.

Thought: ...
Action: ...
\end{PromptBox}
\end{center}
\PromptCaption{fig:prompt-redna-user}{Prompt 11. ReDNA User Prompt Template (ReDNA).}

\end{document}